\pdfoutput=1

\documentclass[11pt]{article}

\usepackage[final]{coling}

\usepackage{times}
\usepackage{latexsym}

\usepackage[T1]{fontenc}

\usepackage[utf8]{inputenc}

\usepackage{microtype}

\usepackage{inconsolata}

\usepackage{graphicx}

\usepackage{multirow}
\usepackage{array}
\usepackage[]{xcolor}
\usepackage{booktabs}
\usepackage{colortbl}
\usepackage{collcell}
\usepackage{calc}
\usepackage{caption}
\usepackage{rotating}
\usepackage{makecell}
\usepackage{tabu}
\usepackage{hhline}
\usepackage{subcaption}
\usepackage{arydshln}
\usepackage{graphicx}
\usepackage{courier}
\usepackage{comment}
\usepackage{enumitem}

\definecolor{stddevcolor}{gray}{0.6}

\newcommand{\pretrainnew}{\texttt{TOP-Training}}
\newcommand{\meqa}{\texttt{Medical-EQA}}

\title{\pretrainnew: Target-Oriented Pretraining for Medical Extractive Question Answering}

\author{
  \textbf{Saptarshi Sengupta \textsuperscript{1}},
  \textbf{Connor Heaton \textsuperscript{1}},
  \textbf{Shreya Ghosh \textsuperscript{2}},\\
  \textbf{Wenpeng Yin \textsuperscript{1}},
  \textbf{Preslav Nakov \textsuperscript{3}},
  \textbf{Suhang Wang\textsuperscript{1}} 
\\
  \textsuperscript{1}Pennsylvania State University, USA \\
  \textsuperscript{2}Indian Institute of Technology (IIT) Bhubaneswar, India \\
  \textsuperscript{3}Mohamed bin Zayed University of Artificial Intelligence, UAE \\
  \{\texttt{sks6765, czh5372, wenpeng, szw494}\}\texttt{@psu.edu} \\ \texttt{shreya@iitbbs.ac.in, preslav.nakov@mbzuai.ac.ae}
}

\begin{document}
\maketitle
\begin{abstract}
We study extractive question-answering in the medical domain (\meqa). This problem has two main challenges: (i) domain specificity, as most AI models lack necessary domain knowledge, and (ii) extraction-based answering style, which restricts most autoregressive LLMs due to potential hallucinations. To handle those challenges, we propose \pretrainnew, a target-oriented pretraining paradigm that stands out among all domain adaptation techniques with two desirable features: (i) \pretrainnew~ moves one step further than popular domain-oriented fine-tuning since it not only moves closer to the target domain, but also familiarizes itself with the target dataset, and (ii) it does not assume the existence of a large set of unlabeled instances from the target domain. Specifically, for a target \meqa~dataset, we extract its entities and leverage large language models (LLMs) to generate synthetic texts containing those entities; we then demonstrate that pretraining on this synthetic text data yields better performance on the target \meqa~ benchmarks. Overall, our contributions are threefold: (i) \pretrainnew, a new pretraining technique to effectively adapt LLMs to better solve a target problem, (ii) \pretrainnew~has a wide application scope because it does not require the target problem to have a large set of unlabeled data, and (iii) our experiments highlight the limitations of autoregressive LLMs, emphasizing \pretrainnew~as a means to unlock the true potential of bidirectional LLMs.\footnote{Our codebase is available \url{https://github.com/saptarshi059/CDQA-v1-Targetted-PreTraining}}
\end{abstract}

\section{Introduction}

The escalating volume of electronic health records (EHR) underscores the growing significance of information extraction (IE) from these datasets. This includes tasks like identifying the medications a patient is currently taking and uncovering recorded drug allergies or adverse reactions. Extractive Question Answering (EQA) plays a central role in EHR-based IE, wherein the system must provide a relevant textual excerpt from medical records based on a query. This is commonly referred to as Medical Extractive Question Answering (\meqa) \cite{10385572}. The primary challenges in \meqa~stem from i) limited data availability, especially expert-labeled data; ii) the presence of rare medical terminologies, which many AI models struggle to recognize and interpret; and iii) privacy concerns that restrict the unfettered use or scraping of open resources.

Despite considerable research efforts and advancements in \meqa, prior work grapples with two main issues: i) Some approaches rely on the assumption of having access to extensive unlabeled data for domain adaptation, which is often impractical due to privacy constraints \cite{10.1145/3531146.3534642}; ii) The recent adoption of decoder-only generative large language models (LLMs) pre-trained on vast datasets holds promise but suffers from \textit{hallucination} issues \cite{ji2023survey, pal-etal-2023-med}, limiting their reliability in medical applications. This raises critical questions: firstly, how can models be trained effectively when the target medical dataset is severely limited, with scarce training data and unavailable unlabeled data? Secondly, how can we harness the rich pre-trained knowledge in LLMs while mitigating hallucination problems?

In this study, we propose \pretrainnew, a target-oriented pretraining technique as a solution to these challenges, addressing them from two angles: i) For target datasets with constrained scales, \pretrainnew~extracts medical entities and prompts LLMs to generate large-scale synthetic data tailored to the target domain, thereby enhancing the model's comprehension of rare medical terminologies; ii) \pretrainnew~reconfigures LLMs as bidirectional encoders to ensure output consistency, alleviating hallucination issues. The fundamental rationale behind \pretrainnew~is to utilize the resource-limited target data as a seed, without accessing gold outputs, and to prompt generative LLMs to produce substantial unlabeled data, thus bolstering extractive LLMs for improved \meqa~ performance.

We evaluate \pretrainnew~ on two prototypical \meqa~ datasets: COVID-QA \cite{moller-etal-2020-covid} and RadQA \cite{soni-etal-2022-radqa}. Our experiments reveal that: i) generative LLMs underperform significantly on these datasets; ii) \pretrainnew~ achieves state-of-the-art performance with moderate-sized synthetic data, offering advantages over baselines that rely on billion-level pre-trained data, often inaccessible in practical settings; iii) \pretrainnew~demonstrates robustness across different encoder-based LLMs, such as BERT and RoBERTa, as well as varying sizes and lengths of synthetic data.
In summary, our contributions are threefold:
\begin{itemize}[leftmargin=*]
    \item We propose a novel LLM-based synthetic data generation approach that leverages medical domain-specific entities to bridge knowledge between target data and synthetic data, facilitating a pretraining stage closely aligned with the target problem and boasting broad application potential.
    \item We offer insights into why generative LLMs struggle in \meqa~and why extractive LLMs are better suited for this task.
    \item We present a fresh perspective on domain adaptation, suggesting a redefinition of "domain" to incorporate not only textual genre but also dataset-specific characteristics, thereby providing a more nuanced understanding of adaptation challenges.
\end{itemize}


\begin{figure*}
    \centering
    \includegraphics{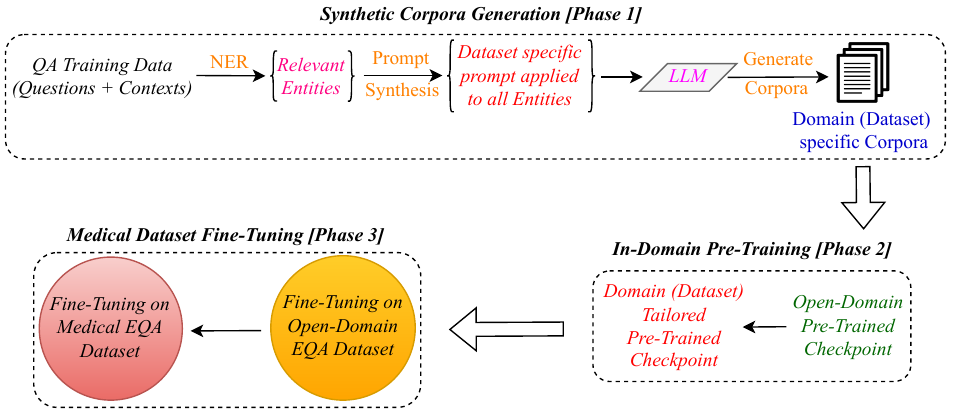}
    \caption{\pretrainnew. First, we extract relevant entities from the target dataset to generate our synthetic pre-training corpora. Next, we train an open-domain model on this corpus followed by two rounds of EQA fine-tuning, i.e., first on an open-domain dataset to learn what EQA is as a task and then on the target \meqa~ dataset.}
    \label{fig:method}
\end{figure*}

\section{Related Work}

In this section, we discuss prior work on domain-specific pre-training, decoder-based Foundation Models (FM) for QA, highlighting their limitations, and, efficient methods for \meqa.

\paragraph{Autoregressive modelling for QA}
\label{subsec:decoder-lms-eqa}

On release, GPT-4 \cite{achiam2023gpt} became \textit{the} benchmark FM on almost all canonical NLP tasks.  
Remarkably, \citet{nori2023can} showed that just by careful prompting, GPT-4 can achieve SOTA on various medical QA datasets. Initiatives such as Meditron-70B \cite{chen2023meditron} 
came next to parallel GPT-4's medical expertise. On closer inspection, however, we see that the datasets on which these two models are applied 
are all multiple-choice style. 
Thus, their capability of handling EQA datasets, especially in the medical domain, remains unexplored (cf. App. \ref{app:chatgpt}).

\citet{xu2021attention} train BART \cite{lewis-etal-2020-bart} for span extraction by considering cross-attention weights as start/end token probabilities to align the generated and true spans while \citet{mallick2023adapting} reframe the problem as generating \textit{numeric indices} indicating either token or sentence-level spans. 
However, we see two issues, i) marginal performance gains, and ii) only \citet{mallick2023adapting} test a \meqa~ dataset, that too without much success as compared to their encoder-based SOTA indicating their ineffectiveness. 
\citet{luo-etal-2022-choose} 
show that \textit{encoders outperform decoders} in span-detection with the added advantage of being better at \textit{out-of-domain generalization}. 
While they do make a case for autoregressive FMs, \citet{liu2023lost} demonstrate how newer instruction-tuned FMs 
are \textit{sensitive to the location of the gold span}. 

\paragraph{Domain Specific Pre-Training}


\citet{gururangan-etal-2020-dont} introduced DAPT (Domain-Adaptive Pretraining) and TAPT (Task-Adaptive Pretraining), which share similarities with our work. DAPT involves extended pretraining on domain-specific corpora without labels, while TAPT focuses on pretraining on the unlabelled training set of the downstream task. Although they demonstrate the effectiveness of TAPT compared to DAPT, closed-domain datasets like COVID-QA typically lack a separate unlabelled training set and may not even have train/dev/test splits. Further, DAPT considers knowledge beyond what is necessary to the task data, whereas our approach confines training to only required concepts.

DAPT/TAPT requires high quantities of unlabelled corpora to yield useful results, thus raising the question: \textit{What happens when we do not have enough ``relevant'' domain data, either in style or volume}? This inspires \pretrainnew, which focuses on a subset of a domain, tailor-made for the downstream dataset, similar in motivation with \citet{gunasekar2023textbooks, zhou2023lima}, who make use of smaller yet \textit{better quality} corpora. 

\paragraph{Efficient Methods in \meqa}
\label{sec:Efficient_methods_in_Medical_EQA} 

We identify seven relevant works on COVID-QA. First off, \citet{haddouche2023transformer} perform a simple evaluation of BERT and RoBERTa and achieve scores similar to ours, thus serving as a sanity check. \citet{zafar2024kimedqa, zafar2024my, poerner-etal-2020-inexpensive} incorporate medical information either via knowledge graphs or training external embeddings on biomedical corpora ranging from 2 to 94 GB. While the former method is ineffective (as showcased by the results), the latter uses much more resources than our method and is still inferior to us. \citet{levy2021open} create a pipeline to provide high-quality references for user queries and shows strong performance on COVID-QA. 
However, there is a caveat to these results. Apart from \citet{poerner-etal-2020-inexpensive}, \citet{zafar2024kimedqa, zafar2024my, levy2021open} all use a \textbf{subset of COVID-QA}. 
\citet{samuel2023can} train on synthetic data from GPT-4 but show weak overall performance as GPT might be unable to create text in the style of COVID-QA. Finally, \citet{seo2024retrieval} generates training data by retrieving samples from a related dataset and prompting an LLM using the original and retrieved data. Although they use fewer samples than us, they generate questions, contexts and answers while we only generate contexts leading to an overall simpler framework capable of outperforming theirs.

For RadQA, we identify three relevant papers. \citet{ghosh-etal-2023-radling} train \texttt{RadLing}, an ELECTRA \cite{clark2020electra} model on a collection of 500K radiology reports which we outperform using far less data (c.f. Table \ref{tab:Rad-QA}). \citet{hernandez2023we} conduct a comprehensive analysis of clinical FMs in the radiology domain 
and show SOTA performance by BioClinicalRoBERTa \cite{lewis-etal-2020-pretrained} once more reinforcing encoder models superiority to causal models for EQA. Finally, \citet{lu2024encode} propose a method to encode input text once and have it be shared across each decoder prompt and apply their method to RadQA to obtain similar EM as the best T5 model shown in \citet{hernandez2023we}. Both \citet{lu2024encode} and \citet{hernandez2023we} show strong performance using Clinical-T5\textsubscript{LARGE} which uses much more data and is a larger model than ours. Despite this, we show how competitive our method is at a fraction of the cost.

\section{Formulation of \meqa}
\label{sec:Formulation_of_medical_EQA}

Each labelled data point consists of three elements, I) \textsc{Context ($\mathcal{C}$)}: A piece of text in the medical domain that introduces the necessary information about a topic; II) \textsc{Question ($\mathcal{Q}$)}: A question sentence acquiring the information from \textsc{Context}. \emph{It can be answerable or unanswerable} and III) \textsc{Answer ($\mathcal{A}$)}: A consecutive span in \textsc{Context} that acts as the answer of \textsc{Question}.

Typical datasets for \meqa~ are split into $train$, $test$ and optional $dev$ sets. There are two main challenges for \meqa~ i) domain specificity; most AI models lack the necessary knowledge; ii) extraction-based answering style, which restricts most autoregressive LLMs due to hallucination concerns. These challenges motivate us to propose \pretrainnew~ accompanied with bi-directional LLMs for this particular problem.

\section{\pretrainnew}
\label{sec:Method}

\pretrainnew, shown in Figure \ref{fig:method}, first extracts entities from the target \meqa~ dataset, then leverages an existing LLM to generate entity-related \textsc{Context} (Section \ref{sec:datageneration}), which acts as the data for further tuning a bi-directional LLM for the target problem (Section \ref{sec:pretrain}).

\subsection{LLM generates target-oriented \textsc{Context}}\label{sec:datageneration}

\paragraph{Entity collection from the target problem.} In this work, we use entities in the target dataset as the connection between it with the newly generated synthetic data. To extract entities from the target EQA dataset, first, we combine all the \textsc{Questions} and \textsc{Contexts} from the $train$ split of the \meqa~ dataset. Next, we extract entities through Named Entity Recognition (NER) using spaCy\footnote{With ``\texttt{en\_core\_sci\_sm}'' (\url{https://allenai.github.io/scispacy/}).}. This step identifies roughly 47k and 11k entities in COVID-QA and RadQA. These entities are aligned with the medical space such as \texttt{DC-SIGNR}, ~\texttt{MTCT},~ \texttt{C-terminal domain}, etc. which are quite different from general-domain entities (person/place/thing) such as \textit{New York}, \textit{John Doe}, etc.


\paragraph{Synthetic \textsc{Context} generation.}
Next, we create prompts for the identified entities to generate \textsc{Contexts} mimicking the target datasets. This required studying the characteristics of the target datasets such as the text genre (full research articles in COVID-QA and radiology reports in RadQA), lengths, and relevant keywords. Galactica \cite{taylor2022galactica}, a generative LLM pre-trained on a collection of text encompassing research articles, knowledge bases, code and \LaTeX\ markup, is used to generate our synthetic data\footnote{Other generative models such as BLOOM \cite{scao2022bloom} and \href{https://github.com/stanford-crfm/BioMedLM}{PubMedGPT} performed worse in our experiments.}. We choose Galactica over other LLMs for two reasons, a) It possesses a built-in ability to generate research paper-like content (see \ref{fbox: covidqa-prompt}) b) Galactica is a general-purpose science model that allows future work to extend our framework to domains beyond biomedical.

The prompts used to generate aligned data \footnote{See Appendix. \ref{app:corpora_samples}} for each dataset are described below.

\textbullet\enspace \textbf{Prompt for COVID-QA.}
Since COVID-QA comprises research papers, based on this characteristic, we develop the following prompt for Galactica to generate pseudo-research articles based on retrieved entities. Note, \texttt{Title} is the prompt handle/keyword and \texttt{entity} is the entity identified in the above stage.\\

\fbox{%
 \begin{minipage}{18em}
 \label{fbox: covidqa-prompt}   \textit{\enspace\enspace\enspace\enspace\enspace\enspace\enspace\enspace\enspace\enspace\colorbox{red!25}{\texttt{Title}}: \colorbox{blue!25}{\texttt{Entity}}}
 \end{minipage}
}\\

\textbullet\enspace \textbf{Prompt for RadQA.} RadQA's \textsc{Contexts} are redacted radiology reports without any consistent format \cite{hartung2020create}. 
The \emph{Findings} and \emph{Impressions} sections are the most vital in a patient's report (akin to the experiment and results section in a research paper). Inspired, we propose the following prompt:\\

\fbox{%
 \begin{minipage}{18em} 
    \textit{\colorbox{red!25}{\texttt{Patient has}} \colorbox{blue!25}{\texttt{Entity}}. \colorbox{red!25}{\texttt{FINDINGS}} \colorbox{red!25}{\texttt{AND}} \colorbox{red!25}{\texttt{IMPRESSION}}:}
 \end{minipage}
}\\

It is worth noting that \textbf{Galactica had not been trained on radiology reports.
Through this prompt, we synthesize \textit{pseudo-reports} bypassing any privacy concerns}. To maintain size parity between the two target datasets, five \textsc{Contexts} are generated for each entity identified in RadQA, yielding around 55k (11k*5) total \textsc{Contexts}, and one \textsc{Context} for each identified entity in COVID-QA, resulting a set of size 47K. 

\subsection{System pre-training on synthetic data}\label{sec:pretrain}
After generating \textsc{Contexts} for each target data, we perform \pretrainnew~ i.e., extended pre-training of BERT/RoBERTa on our generated corpus. In addition, \pretrainnew~ is followed by two rounds of fine-tuning where a) the model will first be fine-tuned on the SQuAD dataset \cite{rajpurkar2016squad, rajpurkar-etal-2018-know} to learn what EQA is in a general domain and b) it is further fine-tuned on the $train$ of each target data (either COVID-QA or RadQA), to solve the $test$.


\section{Experiments}
\label{sec:Exp}

We focus on two datasets: COVID-QA, comprising 2,019 answerable QA pairs (no train/dev/test splits) sourced from CORD-19~\cite{wang-etal-2020-cord}, and RadQA, consisting of 6,148 QA pairs from radiology reports, with a train/dev/test split of 4,878/656/614. We experiment in two areas: baselines and \pretrainnew, and provide mean and standard deviation scores over three random seeds. 

\subsection{Baselines \& influencing factors}\label{subsec:Benchmarking}


Apart from comparison with existing works (c.f. \ref{sec:Efficient_methods_in_Medical_EQA}), we consider 13 encoder models in total taking into account which model made the most sense to apply to either dataset. 
On COVID-QA, we applied models from checkpoints fine-tuned on SQuAD v1 \cite{rajpurkar2016squad} while RadQA, containing unanswerable questions, was tackled with those fine-tuned on SQuAD v2 \cite{rajpurkar-etal-2018-know}. We begin with SQuAD-trained checkpoints as it has been shown \cite{soni-etal-2022-radqa, castelli2020techqa} that models benefit from a first round of training on SQuAD. For consistency, we use the \textit{cased}, \textit{base} version of each architecture when available. 
We use five-fold cross-validation for fine-tuning and report the results in Table \ref{tab:covidqa-all} (COVID-QA).
Results of models applied to the prescribed splits (RadQA) are presented in Table \ref{tab:Rad-QA}. The metrics used are exact match (EM), a binary measure of whether the prediction and gold-standard spans are identical and F1, the harmonic mean of the number of shared words in the two spans with respect to the number of words in the prediction (precision) and with respect to the number of words in the gold-standard span (recall).

We use Galactica-1.3B for consistency with our corpus generation experiments, MedLLaMA(13B) and MedAlpaca(13B) as strong open-source medical checkpoints. We measure the ability of the three decoder models to generate answers without fine-tuning, considering that decoders do not extract spans, but generate answers, for comparison to the gold-standard spans. Following \citet{yue-etal-2021-contrastive}, each sample is formatted as, \\

\fbox{%
 \begin{minipage}{18em}
    \texttt{Question:<question\_text> Context:<part\_of\_context> Answer:}
 \end{minipage}
}
\\


Due to the large size of COVID-QA contexts, they were segmented as they exceeded the maximum sequence length of each model (2,048 tokens). We report overall EM/F1 on each dataset and average best EM/F1 (parenthesis in Table \ref{tab:covidqa-all}) from each Q+C+A chunk for COVID-QA (N/A for RadQA since the context size was much smaller than the models' maximum input length). 

\paragraph{Corpus Size.} We investigated the impact of synthetic dataset size on downstream performance in COVID-QA. We examined the effects of generating one and 10 contexts per entity.

\paragraph{Context Length.} The average context length for COVID-QA is 6K tokens, and Galactica has a maximum window of 2K, resulting in a misalignment between the synthetic corpus and the target dataset. We cannot increase the context size of Galactica.  Training it from scratch with architectural changes is infeasible for us. Thus, we explore the impact of sequence length in the synthetic corpus by limiting the records to only 1k tokens. We cannot determine if \textit{longer} sequences are \textit{beneficial}; we can see if \textit{shorter} ones are \textit{detrimental}.

\paragraph{Prompting Style.} We explore the use of two different prompts when encouraging Galactica to generate \textit{pseudo} radiology reports - as defined in section \ref{sec:datageneration} which we call \textit{Fancy prompt} and \textit{Normal prompt} as simply ``\texttt{\textcolor{blue}{[entity]}}''.


\paragraph{Human-Generated Contexts.} We establish a \textit{Wikipedia} baseline alongside our domain-specific models to assess the influence of content and text structure during domain adaptation. 
Additionally, \citet{micallef-etal-2022-pre} have shown that mBERTu Wiki, pre-trained on Maltese Wikipedia data, surpassed the performance of mBERT, thereby proving to be a competitive baseline. 
For each entity, we query Wikipedia and retrieve the complete page associated with the top search result. The number of entities available for this baseline is much smaller  than
that in our approach since most of the entities do not exist in Wikipedia due to either being extremely esoteric, e.g., \texttt{pulmonary parenchymal infiltrate} or improperly formed, e.g., \texttt{Bao \&}.

\section{Results \& Discussion}
\label{sec:Results & Discussion}

Results for COVID-QA and RadQA are presented in Table \ref{tab:covidqa-all} and \ref{tab:Rad-QA}. Each table is organized as scores from related work, our benchmarks and \pretrainnew. The last 3 rows of the benchmark section provide the zero-shot performance of our chosen decoder models on each dataset. We do not perform multiple trials here as the extremely poor performance would not benefit from additional runs. 
Overall, we see that MedAlpaca is the best among the three for RadQA and only marginally poorer in terms of F1 for COVID-QA. On COVID-QA, no model generates text w.r.t the gold standard 
and only shows positive F1.





\subsection{COVID-QA}

\begin{table*}[ht]
\centering
\vskip -1em
\scalebox{0.7}{
\begin{tabular}{p{1cm}p{3.5cm}p{3cm}p{3cm}ccc}
\hline
& \textbf{Approach/Model} & \textbf{Training Corpus} & \textbf{Corpus Size} & \textbf{Time\#} & \textbf{EM} & \textbf{F1} \\ \midrule
\multirow{8}{*}{\rotatebox[origin=c]{90}{\textbf{Related Work}}} & \citet{haddouche2023transformer} & \multirow{8}{*}{N/A} & \multirow{8}{*}{N/A}  & \multirow{8}{*}{N/A} & 38.61 & 64.87 \\
& \citet{moller-etal-2020-covid}  & & & &  25.9 & 59.53 \\ 
& \citet{levy2021open}& & & &  39.16 & 72.03 \\ 
& \citet{poerner-etal-2020-inexpensive} & & & &  34.62 & 60.23 \\ 
& \citet{zafar2024kimedqa}  & & & &  31.92 & 59.57 \\
& \citet{zafar2024my}  & & & &  35.3 & 58.64 \\
& \citet{samuel2023can} & & & &  31.9 & 58.66 \\
& \citet{seo2024retrieval} & & & & - & \textbf{67.95}* (68.36) \\ 
\bottomrule
\multirow{13}{*}{\rotatebox[origin=c]{90}{\textbf{Our Benchmarks}}} & BioBERT  &  PubMed & 4.5B words & \multirow{13}{*}{N/A} & $37.62\pm\textcolor{stddevcolor}{0.18}$ & $65.73\pm\textcolor{stddevcolor}{0.35}$ \\
& SciBERT  & Semantic Scholar & 3.2B words & & $37.52\pm\textcolor{stddevcolor}{0.23}$ & $65.58\pm\textcolor{stddevcolor}{0.18}$ \\
& \hspace{3mm} +CORD-19 & \hspace{3mm} + CORD-19 & 3.2B words + 20GB & & $35.61\pm\textcolor{stddevcolor}{0.30}$ & $63.60\pm\textcolor{stddevcolor}{0.59}$ \\ 
& PubMedBERT  & PubMed & 3.1B words / 21GB & & $39.87\pm\textcolor{stddevcolor}{0.74}$ & \cellcolor{blue!10}$68.47\pm\textcolor{stddevcolor}{0.13}$ \\
& BlueBERT $^{1}$ &  PubMed + MIMIC & 4.5B words & & $27.35\pm\textcolor{stddevcolor}{0.30}$ & $52.18\pm\textcolor{stddevcolor}{0.40}$ \\
& CODER $^{2}$  &   UMLS & N/A  & & $39.33\pm\textcolor{stddevcolor}{0.47}$ & $67.01\pm\textcolor{stddevcolor}{0.30}$ \\ 
& Longformer $^{3}$ & General Domain & 6.5B tokens & & $37.79\pm\textcolor{stddevcolor}{0.39}$ & $66.58\pm\textcolor{stddevcolor}{0.22}$ \\
& BigBird $^{4}$ & General Domain &  160GB & & $32.79\pm\textcolor{stddevcolor}{0.13}$ & $60.06\pm\textcolor{stddevcolor}{0.41}$ \\
& LUKE $^{5}$ & Wikipedia & 3.5B words & & \cellcolor{blue!10}$41.01\pm\textcolor{stddevcolor}{0.30}$ & $68.23\pm\textcolor{stddevcolor}{0.19}$ \\
& XLNET  & General Domain & 32.89B words &  & \cellcolor{red!10}$2.45\pm\textcolor{stddevcolor}{0.08}$ & \cellcolor{red!10}$8.64\pm\textcolor{stddevcolor}{0.19}$ \\
& Galactica & c.f. section \ref{sec:Method} & 106B tokens & &  0 (0)	& 5.01 (11.11) \\
& MedLLaMA & Medical Corpora & N/A  & & 0 (0)	& \textbf{5.81 (12.79)} \\
& MedAlpaca & \href{https://github.com/kbressem/medAlpaca/blob/main/DATA\_DESCRIPTION.md\#medical-meadow}{Medical Meadow} & N/A & & \textbf{0.03 (0.2)} & 5.21 (12.73) \\
\bottomrule
\multirow{10}{*}{\rotatebox[origin=c]{90}{\textbf{Proposed Method}}} 
& BERT & \multirow{2}{*}{Vanilla Fine-Tuning} & \multirow{2}{*}{N/A} & \multirow{2}{*}{N/A} &  \cellcolor{red!10}$33.62\pm\textcolor{stddevcolor}{0.59}$	& \cellcolor{red!10}$60.01\pm\textcolor{stddevcolor}{0.36}$ \\
& RoBERTa & &  & & $38.89\pm\textcolor{stddevcolor}{0.52}$	& $67.44\pm\textcolor{stddevcolor}{0.47}$ \\ \cmidrule[\lightrulewidth]{2-7}
& BERT & \multirow{2}{*}{Wikipedia} & \multirow{2}{*}{139.6 MB} & \multirow{2}{*}{$\approx$ 2.5 hrs}  &   $33.95\pm\textcolor{stddevcolor}{0.13}$	& $60.76\pm\textcolor{stddevcolor}{0.78}$	 \\
& RoBERTa &  & & & $40.33\pm\textcolor{stddevcolor}{0.60}$ &	$68.30\pm\textcolor{stddevcolor}{0.54}$  \\ \cmidrule[\lightrulewidth]{2-7}
& BERT & \multirow{2}{*}{Gal(47k)}  & \multirow{2}{*}{67.4 MB} &  \multirow{2}{*}{$\approx$ 6.5 hrs}  & $34.97\pm\textcolor{stddevcolor}{0.18}$ & $62.11\pm\textcolor{stddevcolor}{0.32}$ \\ 
& RoBERTa & &  & & $\textbf{41.51}\pm\textcolor{stddevcolor}{\textbf{0.48}}$	& \cellcolor{blue!10}$\textbf{69.10}\pm\textcolor{stddevcolor}{\textbf{0.27}}$  \\ \cmidrule[\lightrulewidth]{2-7}
& BERT & \multirow{2}{*}{Gal(470k) [10x]} & \multirow{2}{*}{558.2 MB} & \multirow{2}{*}{$\approx$ 2.5 days} &  $\textbf{36.39}\pm\textcolor{stddevcolor}{\textbf{0.27}}$	& $\textbf{63.84}\pm\textcolor{stddevcolor}{\textbf{1.16}}$ \\ 
& RoBERTa & &  & & $41.31\pm\textcolor{stddevcolor}{0.22}$	& $68.84\pm\textcolor{stddevcolor}{0.28}$  \\ \cmidrule[\lightrulewidth]{2-7}

& BERT & \multirow{2}{*}{Gal(25k*2 = 50k)$\clubsuit$}  & \multirow{2}{*}{64.0 MB}  & \multirow{2}{*}{$\approx$ 6.5 hrs} & $35.03\pm\textcolor{stddevcolor}{0.38}$ &	$62.14\pm\textcolor{stddevcolor}{0.48}$ \\
& RoBERTa & &  & & $41.36\pm\textcolor{stddevcolor}{0.35}$ & $69.00\pm\textcolor{stddevcolor}{0.52}$  \\ \cmidrule[\lightrulewidth]{2-7}

& BERT & \multirow{2}{*}{Gal(47k)$\spadesuit$}  & \multirow{2}{*}{44.8 MB} & \multirow{2}{*}{$\approx$ 2.5 hrs} & $34.90\pm\textcolor{stddevcolor}{0.14}$ &	$62.02\pm\textcolor{stddevcolor}{0.95}$  \\
& RoBERTa & &  & & \cellcolor{blue!10}$41.57\pm\textcolor{stddevcolor}{0.33}$ & $68.98\pm\textcolor{stddevcolor}{0.31}$ \\ \hdashline
& \textbf{Existing Best} & & & &  41.01 (LUKE) & \textbf{67.95}* (68.36) (\cite{seo2024retrieval}) \\ 
& \textbf{Our Best} & & & &  \textbf{41.51} & \textbf{69.10} \\ 

\bottomrule
\end{tabular}}
\caption{Results for COVID-QA. Top: Scores from Existing Works. *: \citet{seo2024retrieval} claims this score as statistically significant. | Middle: Our Benchmarks. $^{1}$\citet{peng-etal-2019-transfer}; $^{2}$\citet{yuan2022coder}; $^{3}$\citet{beltagy2020longformer}; $^{4}$\citet{zaheer2020big}; $^{5}$\citet{yamada-etal-2020-luke}; \textcolor{blue}{Blue} = \textcolor{blue}{best}/\textcolor{red}{red} = \textcolor{red}{worst} scores overall; \textbf{bold} = \textbf{best decoder} | Bottom: Scores from \pretrainnew. Time\#: Time to generate the corpus; $\clubsuit$: entity filter; $\spadesuit$: maximum context length = 1k tokens; Gal = Galactica; \textcolor{blue}{Blue} = \textcolor{blue}{best}/\textcolor{red}{red} = \textcolor{red}{worst} scores overall; \textbf{bold} = \textbf{best BERT/RoBERTa setup.}} 
\label{tab:covidqa-all}
\end{table*}

We now discuss the results of the benchmarking trials on COVID-QA contrasting the methods.

\subsubsection{Baselines}
\label{subsubsec:covid-benchmark}

Our experiments demonstrate that a one-size-fits-all approach does not work always for domain adaptation. 
BioBERT and PubMedBERT were trained on similar corpora and yet scored in the same range, indicating no clear winner. 
PubMedBERT which is trained from scratch using a custom vocabulary covering a range of medical jargon performs better.

SciBERT (+CORD-19), trained on CORD-19 articles, performs worse than regular SciBERT, suggesting potential issues in training choices or noisiness in the data. Notably, LUKE, trained solely on Wikipedia data, emerges as the best baseline model, possibly due to its entity-recognition pre-training objective, which aids in identifying relevant entities for QA tasks~\cite{van2019does} and highlighting the need for entity representations in closed domains. Models capable of handling longer context, i.e., Longformer, BigBird and XLNet do not show marked improvements. XLNet degrades completely on COVID-QA potentially due to token permutation hindering its reasoning across large and conceptually dense contexts.


Concerning prior work, we analyze \citet{poerner-etal-2020-inexpensive} and \citet{seo2024retrieval} as these are the only two directly comparable works. \citet{poerner-etal-2020-inexpensive} equips BERT with Word2Vec embeddings trained on corpora ranging from 2-94 GB whereas we use a 67.4 MB corpus and obtain slightly better EM and $\sim$3\% more F1 (BERT: proposed method, row 3, Table \ref{tab:covidqa-all}). Additionally, RoBERTa with \pretrainnew ~shows 1.7\% more F1 (proposed method, row 3, Table \ref{tab:covidqa-all}) compared to \citet{seo2024retrieval} who use a much larger model (T5) for their approach. Both instances show the resource benefits coming from ~\pretrainnew.

\subsubsection{\pretrainnew}

\textbf{Wiki Baseline}: Fine-Tuning BERT on Wikipedia yields marginal improvement of 0.9\% EM and 1.2\% F1; RoBERTa shows a 3.7\% increase in EM and a 1.3\% increase in F1. Our Wikipedia corpus, while small, contains relevant information about COVID literature, which in turn aids in answering related questions. 

\textbf{47k corpus}: With \pretrainnew, BERT achieves a 4.01\% increase in EM and a 3.5\% increase in F1, while RoBERTa shows a 6.7\% increase in EM and a 2.5\% increase in F1, {\bf setting a new SOTA on COVID-QA}. RoBERTa even outperforms the previous SOTA model (LUKE) by 1.2\% in EM and 1.3\% in F1, despite using a training corpus significantly smaller (67.4 MB/0.032B words) than LUKE's 3.5B-word corpus (0.9\% of the size). Moreover, any variation of our approach, ablation or otherwise, improves performance for both models over the Wikipedia baseline. These results indicate that our models benefit from additional training on corpora aligned with entity information from the downstream dataset.

\textbf{470k [10x] corpus}: Training with a 10x corpus (10 contexts per entity) led to the most improvements for BERT with EM increasing by 8.2\% and F1 by 6.4\%. This is consistent with \citet{liu2019roberta} who argue that BERT was \textit{significantly under-trained}. Though this improvement does not achieve RoBERTa performance, it demonstrates the scalability of our approach. 
RoBERTa improves too, but not as much as when using the base 47k corpus as compared to BERT, which, being under-trained, is more \textit{malleable} to learning new concepts while RoBERTa seems to have hit its \textit{ceiling for learning} in this domain.

\textbf{Filtration - 50k corpus}:
Surprisingly, on removing ill-formed entities (proposed method, row 5, Table \ref{tab:covidqa-all}) the performance declined with respect to the best BERT (proposed method, 10x corpus - row 4) and best RoBERTa (proposed method, base 47k corpus - row 3) model. We attribute this to our regex rules which mistakenly (as they cannot distinguish between true/false patterns) removed entities relevant to research articles such as author names or URLs, leading to the decline in performance.

\textbf{Reduced context length - 47k (at most 1k context tokens) corpus}: Galactica can process at most 2048 input tokens which meant that we could not generate text beyond this limit. Instead, we wanted to see if a lower number of tokens, i.e., smaller context size impacted performance (last row of the proposed method, Table \ref{tab:covidqa-all}). Inevitably, both models perform worse on both metrics as Galactica is unable to generate content matching the style of COVID-QA contexts, underscoring the importance of domain-aware writing styles for adaptation pipelines.


\subsection{RadQA}

\begin{table*}
\centering
\setlength{\extrarowheight}{0pt}
\addtolength{\extrarowheight}{\aboverulesep}
\addtolength{\extrarowheight}{\belowrulesep}
\setlength{\aboverulesep}{0pt}
\setlength{\belowrulesep}{0pt}
\LARGE
\resizebox{\textwidth}{!}{
\begin{tabular}{llllcccclcccc}
\hline
 & \textbf{Model} & \textbf{Train Dataset / Corpus Size} & \textbf{Time\#} & \multicolumn{4}{c}{\textbf{Dev}} & \multicolumn{1}{c}{} & \multicolumn{4}{c}{\textbf{Test}} \\ 
\cmidrule(l){5-8}\cmidrule{10-13}
 &  &  &  & \textbf{EM} & \textbf{F1} & \textbf{H(EM)} & \textbf{H(F1)} &  & \textbf{EM} & \textbf{F1} & \textbf{H(EM)} & \textbf{H(F1)} \\ 
\cline{2-8}\cline{10-13}
\multirow{3}{*}{\rotcell{\begin{tabular}[c]{@{}l@{}}\textbf{Related }\\\textbf{Work}\end{tabular}}} & \citet{ghosh-etal-2023-radling} & \multirow{4}{*}{N/A} & \multirow{4}{*}{N/A} &  &  &  &  &  & – & 62.55 &  &  \\
& \citet{lu2024encode} &  &  &  &  &  &  &  & 54.6 & – &  &  \\
& \citet{hernandez2023we} &  &  &  &  &  &  &  & 55.0 & 74.5 &  &  \\ 
& \citet{lewis-etal-2020-pretrained} &  &  &  &  &  &  &  & 60.4 & 75.9 &  &  \\ 
 \cdashline{2-13}
 & \textbf{\pretrainnew} ~ (best) &  &  &  &  &  &  &  & 52.39 & 65.57 & & \\
\bottomrule
\multirow{12}{*}{\rotcell{\begin{tabular}[c]{@{}l@{}}\textbf{Our}\\\textbf{Benchmarks}\end{tabular}}} & BioBERT & PubMed/4.5B words & \multirow{12}{*}{N/A} & {\cellcolor[rgb]{1,0.902,0.902}}\cellcolor[rgb]{1,0.902,0.902}\cellcolor[rgb]{1,0.902,0.902}\cellcolor{red!10}$26.42\pm\textcolor{stddevcolor}{0.49}$ & {\cellcolor[rgb]{1,0.902,0.902}}\cellcolor[rgb]{1,0.902,0.902}\cellcolor[rgb]{1,0.902,0.902}\cellcolor{red!10}$44.26\pm\textcolor{stddevcolor}{0.09}$ & {\cellcolor[rgb]{1,0.902,0.902}}\cellcolor[rgb]{1,0.902,0.902}\cellcolor[rgb]{1,0.902,0.902}\cellcolor{red!10}$40.79\pm\textcolor{stddevcolor}{0.76}$ & {\cellcolor[rgb]{1,0.902,0.902}}\cellcolor[rgb]{1,0.902,0.902}\cellcolor[rgb]{1,0.902,0.902}\cellcolor{red!10}$68.31\pm\textcolor{stddevcolor}{0.14}$ &  & {\cellcolor[rgb]{1,0.902,0.902}}\cellcolor[rgb]{1,0.902,0.902}\cellcolor[rgb]{1,0.902,0.902}\cellcolor{red!10}$49.95\pm\textcolor{stddevcolor}{1.08}$ & {\cellcolor[rgb]{1,0.902,0.902}}\cellcolor[rgb]{1,0.902,0.902}\cellcolor[rgb]{1,0.902,0.902}\cellcolor{red!10}$63.32\pm\textcolor{stddevcolor}{0.40}$ & {\cellcolor[rgb]{1,0.902,0.902}}\cellcolor[rgb]{1,0.902,0.902}\cellcolor[rgb]{1,0.902,0.902}\cellcolor{red!10}$45.65\pm\textcolor{stddevcolor}{1.21}$ & {\cellcolor[rgb]{1,0.902,0.902}}\cellcolor[rgb]{1,0.902,0.902}\cellcolor[rgb]{1,0.902,0.902}\cellcolor{red!10}$63.50\pm\textcolor{stddevcolor}{0.57}$ \\
 & SciBERT & Semantic Scholar/3.2B words &  & $27.03\pm\textcolor{stddevcolor}{0.32}$ & $44.40\pm\textcolor{stddevcolor}{0.06}$ & $41.65\pm\textcolor{stddevcolor}{0.62}$ & $68.45\pm\textcolor{stddevcolor}{0.22}$ &  & $53.04\pm\textcolor{stddevcolor}{0.38}$ & $67.17\pm\textcolor{stddevcolor}{0.73}$ & $48.62\pm\textcolor{stddevcolor}{0.70}$ & $67.49\pm\textcolor{stddevcolor}{1.02}$ \\
 & PubMedBERT & PubMed/3.1B words(21GB) &  & $31.45\pm\textcolor{stddevcolor}{0.17}$ & $47.89\pm\textcolor{stddevcolor}{0.46}$ & {\cellcolor[rgb]{0.902,0.902,1}}\cellcolor[rgb]{0.902,0.902,1}\cellcolor[rgb]{0.902,0.902,1}\cellcolor{blue!10} $48.40\pm\textcolor{stddevcolor}{0.27}$ & {\cellcolor[rgb]{0.902,0.902,1}}\cellcolor[rgb]{0.902,0.902,1}\cellcolor[rgb]{0.902,0.902,1}\cellcolor{blue!10}$73.77\pm\textcolor{stddevcolor}{0.62}$ &  & $54.07\pm\textcolor{stddevcolor}{0.71}$ & {\cellcolor[rgb]{0.902,0.902,1}}\cellcolor[rgb]{0.902,0.902,1}\cellcolor[rgb]{0.902,0.902,1}\cellcolor{blue!10}$68.76\pm\textcolor{stddevcolor}{0.22}$ & $49.49\pm\textcolor{stddevcolor}{0.87}$ & $69.09\pm\textcolor{stddevcolor}{0.80}$ \\
 & BlueBERT & PubMed + MIMIC/4.5B words &  & $30.08\pm\textcolor{stddevcolor}{1.33}$ & $47.14\pm\textcolor{stddevcolor}{0.81}$ & $46.12\pm\textcolor{stddevcolor}{2.01}$ & $73.44\pm\textcolor{stddevcolor}{2.78}$ &  & {\cellcolor[rgb]{0.902,0.902,1}}\cellcolor[rgb]{0.902,0.902,1}\cellcolor[rgb]{0.902,0.902,1}\cellcolor{blue!10}$54.99\pm\textcolor{stddevcolor}{1.91}$ & $68.11\pm\textcolor{stddevcolor}{1.41}$ & $48.55\pm\textcolor{stddevcolor}{1.66}$ & $66.06\pm\textcolor{stddevcolor}{1.40}$ \\
 & CODER & UMLS/N/A &  & {\cellcolor[rgb]{0.902,0.902,1}}\cellcolor[rgb]{0.902,0.902,1}\cellcolor[rgb]{0.902,0.902,1}\cellcolor{blue!10}$40.50\pm\textcolor{stddevcolor}{1.31}$ & {\cellcolor[rgb]{0.902,0.902,1}}\cellcolor[rgb]{0.902,0.902,1}\cellcolor[rgb]{0.902,0.902,1}\cellcolor{blue!10}$57.32\pm\textcolor{stddevcolor}{1.74}$ & $47.37\pm\textcolor{stddevcolor}{1.70}$ & $73.34\pm\textcolor{stddevcolor}{1.29}$ &  & $53.74\pm\textcolor{stddevcolor}{0.71}$ & $68.36\pm\textcolor{stddevcolor}{0.36}$ & $49.86\pm\textcolor{stddevcolor}{0.50}$ & {\cellcolor[rgb]{0.902,0.902,1}}\cellcolor[rgb]{0.902,0.902,1}\cellcolor[rgb]{0.902,0.902,1}\cellcolor{blue!10}$69.36\pm\textcolor{stddevcolor}{1.00}$ \\
 & LUKE & Wikipedia/3.5B words &  & $27.44\pm\textcolor{stddevcolor}{0.70}$ & $44.77\pm\textcolor{stddevcolor}{0.40}$ & $42.35\pm\textcolor{stddevcolor}{1.08}$ & $69.10\pm\textcolor{stddevcolor}{0.62}$ &  & $50.92\pm\textcolor{stddevcolor}{1.26}$ & $64.47\pm\textcolor{stddevcolor}{1.75}$ & $46.16\pm\textcolor{stddevcolor}{0.25}$ & $64.25\pm\textcolor{stddevcolor}{1.28}$ \\
 & RadBERT$^1$ & Radiology reports/2.6 GB &  & $30.34\pm\textcolor{stddevcolor}{1.50}$ & $48.00\pm\textcolor{stddevcolor}{1.43}$ & $45.73\pm\textcolor{stddevcolor}{0.68}$ & $73.00\pm\textcolor{stddevcolor}{0.73}$ &  & $54.40\pm\textcolor{stddevcolor}{2.84}$ & $67.34\pm\textcolor{stddevcolor}{1.74}$ & {\cellcolor[rgb]{0.902,0.902,1}}\cellcolor[rgb]{0.902,0.902,1}\cellcolor[rgb]{0.902,0.902,1}\cellcolor{blue!10}$51.52\pm\textcolor{stddevcolor}{0.87}$ & $68.80\pm\textcolor{stddevcolor}{0.80}$ \\
 & ClinicalBERT$^2$ & MIMIC/0.5B words(3.7GB) &  & $27.18\pm\textcolor{stddevcolor}{1.89}$ & $44.69\pm\textcolor{stddevcolor}{0.54}$ & $41.88\pm\textcolor{stddevcolor}{2.86}$ & $68.90\pm\textcolor{stddevcolor}{0.71}$ &  & $50.27\pm\textcolor{stddevcolor}{1.63}$ & $63.40\pm\textcolor{stddevcolor}{1.52}$ & $46.89\pm\textcolor{stddevcolor}{0.13}$ & $64.41\pm\textcolor{stddevcolor}{0.16}$ \\
 & BioMed-RoBERTa$^3$ & S2ORC/7.55B tokens(47GB) &  & $27.44\pm\textcolor{stddevcolor}{1.10}$ & $45.44\pm\textcolor{stddevcolor}{0.64}$ & $42.35\pm\textcolor{stddevcolor}{1.70}$ & $70.14\pm\textcolor{stddevcolor}{0.99}$ &  & $52.82\pm\textcolor{stddevcolor}{0.57}$ & $66.52\pm\textcolor{stddevcolor}{0.32}$ & $48.62\pm\textcolor{stddevcolor}{0.33}$ & $66.91\pm\textcolor{stddevcolor}{0.82}$ \\
 & Galactica & c.f. section \ref{sec:Method}/106B tokens &  & 1.37 & 8.5 & 1.37 & 8.5 &  & 0.49 & 10.23 & 0.49 & 10.23 \\
 & MedLLaMA & Medical Corpora/ N/A &  & 0.3 & 10.63 & 0.3 & 10.63 &  & 0.16 & 12.14 & 0.16 & 12.14 \\
 & MedAlpaca & \href{https://github.com/kbressem/medAlpaca/blob/main/DATA\_DESCRIPTION.md\#medical-meadow}{Medical Meadow}/ N/A &  & \textbf{1.68} & \textbf{15.18} & \textbf{1.68} & \textbf{15.18} &  & \textbf{1.3} & \textbf{16.95} & \textbf{1.3} & \textbf{16.95} \\ 
\bottomrule
\multirow{16}{*}{\rotcell{\begin{tabular}[c]{@{}l@{}}\textbf{Proposed }\\\textbf{Method}\end{tabular}}} & BERT & \multirow{2}{*}{N/A*} & \multirow{2}{*}{N/A} & {\cellcolor[rgb]{1,0.902,0.902}}\cellcolor[rgb]{1,0.902,0.902}\cellcolor[rgb]{1,0.902,0.902}\cellcolor{red!10}$23.83\pm\textcolor{stddevcolor}{0.49}$ & $42.91\pm\textcolor{stddevcolor}{0.46}$ & {\cellcolor[rgb]{1,0.902,0.902}}\cellcolor[rgb]{1,0.902,0.902}\cellcolor[rgb]{1,0.902,0.902}\cellcolor{red!10}$36.79\pm\textcolor{stddevcolor}{0.76}$ & $66.23\pm\textcolor{stddevcolor}{0.70}$ &  & {\cellcolor[rgb]{1,0.902,0.902}}\cellcolor[rgb]{1,0.902,0.902}\cellcolor[rgb]{1,0.902,0.902}\cellcolor{red!10}$46.20\pm\textcolor{stddevcolor}{1.96}$ & $59.42\pm\textcolor{stddevcolor}{1.10}$ & $40.65\pm\textcolor{stddevcolor}{1.15}$ & $58.30\pm\textcolor{stddevcolor}{0.74}$ \\
 & RoBERTa &  &  & $26.12\pm\textcolor{stddevcolor}{0.69}$ & $43.83\pm\textcolor{stddevcolor}{0.44}$ & $40.31\pm\textcolor{stddevcolor}{1.06}$ & $67.65\pm\textcolor{stddevcolor}{0.68}$ &  & $51.68\pm\textcolor{stddevcolor}{0.73}$ & $64.94\pm\textcolor{stddevcolor}{0.62}$ & $46.45\pm\textcolor{stddevcolor}{0.66}$ & $64.14\pm\textcolor{stddevcolor}{0.42}$ \\ 
\hhline{~-------~----}
 & BERT & \multirow{2}{*}{Wikipedia/18.4 MB} & \multirow{2}{*}{$\approx$30 mins} & $24.49\pm\textcolor{stddevcolor}{0.38}$ & {\cellcolor[rgb]{1,0.902,0.902}}\cellcolor[rgb]{1,0.902,0.902}\cellcolor[rgb]{1,0.902,0.902}\cellcolor{red!10}$42.62\pm\textcolor{stddevcolor}{0.08}$ & $37.80\pm\textcolor{stddevcolor}{0.59}$ & $65.79\pm\textcolor{stddevcolor}{0.13}$ &  & $47.40\pm\textcolor{stddevcolor}{1.85}$ & $60.11\pm\textcolor{stddevcolor}{1.59}$ & $41.95\pm\textcolor{stddevcolor}{1.73}$ & $58.92\pm\textcolor{stddevcolor}{1.56}$ \\
 & RoBERTa &  &  & $27.19\pm\textcolor{stddevcolor}{0.32}$ & $44.49\pm\textcolor{stddevcolor}{0.31}$ & $41.96\pm\textcolor{stddevcolor}{0.49}$ & $68.68\pm\textcolor{stddevcolor}{0.47}$ &  & $50.54\pm\textcolor{stddevcolor}{0.62}$ & $63.43\pm\textcolor{stddevcolor}{0.75}$ & $45.72\pm\textcolor{stddevcolor}{0.50}$ & $62.92\pm\textcolor{stddevcolor}{0.67}$ \\ 
\cline{2-13}
 & BERT & \multirow{2}{*}{Gal($\approx$55k) $\dagger$/ 81.6 MB} & \multirow{2}{*}{$\approx$11 hrs} & $24.70\pm\textcolor{stddevcolor}{0.46}$ & $42.88\pm\textcolor{stddevcolor}{0.30}$ & $38.12\pm\textcolor{stddevcolor}{0.71}$ & $66.19\pm\textcolor{stddevcolor}{0.46}$ &  & $46.74\pm\textcolor{stddevcolor}{1.84}$ & $59.69\pm\textcolor{stddevcolor}{0.62}$ & $41.60\pm\textcolor{stddevcolor}{2.31}$ & $58.88\pm\textcolor{stddevcolor}{1.64}$ \\
 & RoBERTa &  &  & $27.09\pm\textcolor{stddevcolor}{0.70}$ & $44.42\pm\textcolor{stddevcolor}{0.55}$ & $41.81\pm\textcolor{stddevcolor}{1.09}$ & $68.56\pm\textcolor{stddevcolor}{0.86}$ &  & $51.25\pm\textcolor{stddevcolor}{0.41}$ & $64.41\pm\textcolor{stddevcolor}{0.95}$ & $46.45\pm\textcolor{stddevcolor}{0.66}$ & $64.01\pm\textcolor{stddevcolor}{0.71}$ \\ 
\cdashline{2-13}

 & BERT & \multirow{2}{*}{Gal($\approx$55k) $\dagger \clubsuit$/ 80.3 MB} & \multirow{2}{*}{$\approx$11 hrs} & $25.51\pm\textcolor{stddevcolor}{0.38}$ & $43.11\pm\textcolor{stddevcolor}{0.23}$ & $40.38\pm\textcolor{stddevcolor}{2.31}$ & {\cellcolor[rgb]{1,0.902,0.902}}\cellcolor[rgb]{1,0.902,0.902}\cellcolor[rgb]{1,0.902,0.902}\cellcolor{red!10}$64.76\pm\textcolor{stddevcolor}{2.75}$ &  & $46.36\pm\textcolor{stddevcolor}{1.79}$ & $59.41\pm\textcolor{stddevcolor}{1.67}$ & $46.59\pm\textcolor{stddevcolor}{8.06}$ & $57.83\pm\textcolor{stddevcolor}{1.81}$ \\
 & RoBERTa &  &  & $27.69\pm\textcolor{stddevcolor}{0.23}$ & {\cellcolor[rgb]{0.902,0.902,1}}\cellcolor[rgb]{0.902,0.902,1}\cellcolor[rgb]{0.902,0.902,1}\cellcolor{blue!10}$45.16\pm\textcolor{stddevcolor}{0.35}$ & $42.74\pm\textcolor{stddevcolor}{0.36}$ & $69.70\pm\textcolor{stddevcolor}{0.54}$ &  & $51.14\pm\textcolor{stddevcolor}{0.49}$ & $64.29\pm\textcolor{stddevcolor}{0.38}$ & $46.31\pm\textcolor{stddevcolor}{1.36}$ & $63.85\pm\textcolor{stddevcolor}{0.76}$ \\ 
\cline{2-13}
 & BERT & \multirow{2}{*}{Gal($\approx$55k) $\ddagger$/ 38.1 MB} & \multirow{2}{*}{$\approx$11 hrs} & $25.10\pm\textcolor{stddevcolor}{0.32}$ & $42.78\pm\textcolor{stddevcolor}{0.55}$ & $38.74\pm\textcolor{stddevcolor}{0.49}$ & $66.03\pm\textcolor{stddevcolor}{0.84}$ &  & $46.85\pm\textcolor{stddevcolor}{1.74}$ & $59.54\pm\textcolor{stddevcolor}{1.19}$ & $41.66\pm\textcolor{stddevcolor}{0.45}$ & $58.60\pm\textcolor{stddevcolor}{0.37}$ \\
 & RoBERTa &  &  & $27.64\pm\textcolor{stddevcolor}{0.49}$ & $44.99\pm\textcolor{stddevcolor}{0.11}$ & $42.67\pm\textcolor{stddevcolor}{0.76}$ & $69.45\pm\textcolor{stddevcolor}{0.16}$ &  & {\cellcolor[rgb]{0.902,0.902,1}}\cellcolor[rgb]{0.902,0.902,1}\cellcolor[rgb]{0.902,0.902,1}\cellcolor{blue!10}$52.39\pm\textcolor{stddevcolor}{0.80}$ & {\cellcolor[rgb]{0.902,0.902,1}}\cellcolor[rgb]{0.902,0.902,1}\cellcolor[rgb]{0.902,0.902,1}\cellcolor{blue!10}$65.57\pm\textcolor{stddevcolor}{0.93}$ & {\cellcolor[rgb]{0.902,0.902,1}}\cellcolor[rgb]{0.902,0.902,1}\cellcolor[rgb]{0.902,0.902,1}\cellcolor{blue!10}$47.76\pm\textcolor{stddevcolor}{1.85}$ & {\cellcolor[rgb]{0.902,0.902,1}}\cellcolor[rgb]{0.902,0.902,1}\cellcolor[rgb]{0.902,0.902,1}\cellcolor{blue!10}$65.34\pm\textcolor{stddevcolor}{1.97}$ \\ 
\cdashline{2-13}
 & BERT & \multirow{2}{*}{Gal($\approx$55k) $\ddagger \clubsuit$/ 34.3 MB} & \multirow{2}{*}{$\approx$11 hrs} & $\textbf{25.76}\pm\textcolor{stddevcolor}{\textbf{0.66}}$ & $\textbf{43.10}\pm\textcolor{stddevcolor}{\textbf{0.27}}$ & $\textbf{39.68}\pm\textcolor{stddevcolor}{\textbf{1.11}}$ & $\textbf{66.44}\pm\textcolor{stddevcolor}{\textbf{0.55}}$ &  & $46.52\pm\textcolor{stddevcolor}{1.20}$ & {\cellcolor[rgb]{1,0.902,0.902}}\cellcolor[rgb]{1,0.902,0.902}\cellcolor[rgb]{1,0.902,0.902}\cellcolor{red!10}$58.98\pm\textcolor{stddevcolor}{1.09}$ & {\cellcolor[rgb]{1,0.902,0.902}}\cellcolor[rgb]{1,0.902,0.902}\cellcolor[rgb]{1,0.902,0.902}\cellcolor{red!10}$40.44\pm\textcolor{stddevcolor}{0.95}$ & {\cellcolor[rgb]{1,0.902,0.902}}\cellcolor[rgb]{1,0.902,0.902}\cellcolor[rgb]{1,0.902,0.902}\cellcolor{red!10}$57.06\pm\textcolor{stddevcolor}{0.94}$ \\
 & RoBERTa &  &  & $27.08\pm\textcolor{stddevcolor}{0.46}$ & $44.67\pm\textcolor{stddevcolor}{0.23}$ & $41.81\pm\textcolor{stddevcolor}{0.72}$ & $68.95\pm\textcolor{stddevcolor}{0.35}$ &  & $51.30\pm\textcolor{stddevcolor}{1.17}$ & $64.14\pm\textcolor{stddevcolor}{0.41}$ & $47.39\pm\textcolor{stddevcolor}{1.64}$ & $64.53\pm\textcolor{stddevcolor}{0.85}$ \\ 
\cline{2-13}
 & BERT & \multirow{2}{*}{Gal($\approx$100k) $\dagger \ddagger$/ 120.8 MB} & \multirow{2}{*}{$\approx$22 hrs} & $25.10\pm\textcolor{stddevcolor}{0.54}$ & $42.74\pm\textcolor{stddevcolor}{0.06}$ & $38.74\pm\textcolor{stddevcolor}{0.83}$ & $65.97\pm\textcolor{stddevcolor}{0.09}$ &  & $47.07\pm\textcolor{stddevcolor}{1.23}$ & $60.04\pm\textcolor{stddevcolor}{1.19}$ & $42.46\pm\textcolor{stddevcolor}{0.13}$ & $59.78\pm\textcolor{stddevcolor}{0.41}$ \\
 & RoBERTa &  &  & $27.49\pm\textcolor{stddevcolor}{0.49}$ & $44.82\pm\textcolor{stddevcolor}{0.38}$ & $42.43\pm\textcolor{stddevcolor}{0.75}$ & $69.17\pm\textcolor{stddevcolor}{0.58}$ &  & $51.79 \pm\textcolor{stddevcolor}{1.98}$ & $64.99\pm\textcolor{stddevcolor}{1.80}$ & $47.03\pm\textcolor{stddevcolor}{1.45}$ & $64.45\pm\textcolor{stddevcolor}{1.21}$ \\ 
\cdashline{2-13}
 & BERT & \multirow{2}{*}{Gal($\approx$100k) $\dagger \ddagger \clubsuit$/ 115.6 MB} & \multirow{2}{*}{$\approx$22 hrs} & $24.75\pm\textcolor{stddevcolor}{0.23}$ & $42.96\pm\textcolor{stddevcolor}{0.10}$ & $38.20\pm\textcolor{stddevcolor}{0.36}$ & $66.31\pm\textcolor{stddevcolor}{0.16}$ &  & $47.50\pm\textcolor{stddevcolor}{1.06}$ & $60.38\pm\textcolor{stddevcolor}{0.48}$ & $41.96\pm\textcolor{stddevcolor}{1.31}$ & $59.14\pm\textcolor{stddevcolor}{1.97}$ \\
 & RoBERTa &  &  & {\cellcolor[rgb]{0.902,0.902,1}}\cellcolor[rgb]{0.902,0.902,1}\cellcolor[rgb]{0.902,0.902,1}\cellcolor{blue!10}$\textbf{27.85}\pm\textcolor{stddevcolor}{\textbf{0.09}}$ & {\cellcolor[rgb]{0.902,0.902,1}}\cellcolor[rgb]{0.902,0.902,1}\cellcolor[rgb]{0.902,0.902,1}\cellcolor{blue!10}$\textbf{45.16}\pm\textcolor{stddevcolor}{\textbf{0.11}}$ & {\cellcolor[rgb]{0.902,0.902,1}}\cellcolor[rgb]{0.902,0.902,1}\cellcolor[rgb]{0.902,0.902,1}\cellcolor{blue!10}$\textbf{42.98}\pm\textcolor{stddevcolor}{\textbf{0.14}}$ & {\cellcolor[rgb]{0.902,0.902,1}}\cellcolor[rgb]{0.902,0.902,1}\cellcolor[rgb]{0.902,0.902,1}\cellcolor{blue!10}$\textbf{69.71}\pm\textcolor{stddevcolor}{\textbf{0.18}}$ &  & $51.79\pm\textcolor{stddevcolor}{0.59}$ & $64.77\pm\textcolor{stddevcolor}{0.40}$ & $46.88\pm\textcolor{stddevcolor}{1.28}$ & $64.21\pm\textcolor{stddevcolor}{0.75}$ \\
 \bottomrule
\end{tabular}
}
\caption{Results for RadQA. Top: Comparison with Existing Works. | Middle: Our Baselines. H(F1)/H(EM): Has\_F1/Has\_EM i.e F1/EM for answerable questions; \textcolor{blue}{Blue} = \textcolor{blue}{best}/\textcolor{red}{red} = \textcolor{red}{worst} scores overall; \textbf{bold} = \textbf{best decoder} | Bottom: \pretrainnew~Results. *: [Vanilla Fine-Tuning]. [$\dagger$: normal, $\ddagger$: fancy] prompt, $\clubsuit$: entity filter. Time\#: Time to generate the corpus. Gal = Galactica; \textcolor{blue}{Blue} = \textcolor{blue}{best}/\textcolor{red}{red} = \textcolor{red}{worst} overall; \textbf{bold} = \textbf{best BERT/RoBERTa setup}.}
\label{tab:Rad-QA}
\end{table*}

Results on RadQA are presented for both its validation and test splits. We consider various combinations of contexts (prompts) and entity filtration. 
Despite finding no information leakage, higher test scores on average are observed as compared to validation scores, which we attribute to 
fewer unanswerable questions in the test set (154 vs. 231) and slightly shorter contexts (73.82 vs. 78.1 tokens). As such, while we report scores for both, we mainly focus our analysis on the validation set.

\subsubsection{Baselines}
\label{subsubsec:radqa-benchmark}

On the RadQA benchmark dev set, CODER, a PubMEDBERT checkpoint, has the best EM and F1 but suffers slightly v/s PubMedBERT on only answerable questions presumably because CODER learned clinical embeddings from the UMLS knowledge graph with 
radiology terms. 

Surprisingly, PubMed/Blue-BERT performed similarly on both the dev and test sets.
Theoretically, BlueBERT should have performed better being pre-trained on MIMIC clinical notes. RadBERT, which is a superior RoBERTa architecture, and specifically trained on radiology reports did not perform well overall. Although it marginally improved over PubMed/Blue-BERT, it comes at the cost of a fraction of the training data. This again indicates the importance of proper domain alignment, i.e., \textit{what} data the models are trained on. 

Unfortunately, LUKE performed poorly as compared to Bio/Sci-BERT, showing little, v/s both on dev, to no gain, v/s SciBERT on test, during evaluation. The impact of writing styles in the training corpora is evident in the performance gap between Clinical/Rad-BERT. While the former was trained on more clinical data, it was not the \textit{right} type of data, i.e., radiology reports, leading RadBERT to outperform it on both splits.

Contrasting \pretrainnew~ with prior work, we see that by using our corpus with fancy prompts on unfiltered entities (proposed method, row 5, Table \ref{tab:Rad-QA}), RoBERTa outperforms RadLing \cite{ghosh-etal-2023-radling} in test F1 by 4.8\%, the latter using a much larger corpus of radiology reports. Although we are unable to reach SOTA on RadQA, our best RoBERTa, on the test set (proposed method, row 5, Table \ref{tab:Rad-QA}) comes within 2 points absolute EM of Clinical-T5\textsubscript{LARGE} \cite{lu2024encode} which again shows the competitiveness of our proposed method as all our models are BASE versions.



\subsubsection{\pretrainnew}


\textbf{Wiki Baseline}: Training on the Wikipedia corpus, BERT shows an increase in EM only (2.8\%) on the dev set, but an overall improvement in all measures on the test set compared to vanilla fine-tuning. RoBERTa showed overall improvement on the dev set while test scores suffered compared to the vanilla baseline.

\textbf{Normal Prompting}: Training on the unfiltered corpus with normal prompts (proposed method, row 3, Table \ref{tab:Rad-QA}) improved BERT and RoBERTa 3.7\% EM, and 1.4\% F1 for RoBERTa, over the vanilla baseline, which is greater than 
that of our Wiki baseline. However, \textbf{when the filter is applied (proposed method, row 4, Table \ref{tab:Rad-QA})} BERT shows $\sim$7\% EM \& 0.5\% F1 increase while RoBERTa shows \~6\% EM \& \~3\% F1 increase over regular fine-tuning indicating the benefits of data filtration. 

\textbf{Best BERT setting}: When both filtered entities and the corpus from the fancy prompt are used (proposed method, row 6, Table \ref{tab:Rad-QA}), improvements over basic fine-tuning (8.1\% EM, 0.4\% F1 on dev) and the Wikipedia baseline (5.2\% EM, 1.1\%F1 on dev) occur. Note: a) the benefit of studying RadQA and designing targeted prompts, and b) that BERT reaches these scores with a modest 34.3 MB corpus, much smaller than the benchmarked models.


\textbf{Best RoBERTa setting}: RoBERTa demonstrates improvements across different combinations of filtration methods, prompt styles and using the Wikipedia corpus. However, the improvements are inconsistent with respect to a specific approach. Excluding the combined corpora settings, we see it achieve the best performance, on validation, using the corpus obtained from the filtered entities and normal prompting (proposed method, row 4, Table \ref{tab:Rad-QA}) and on the test, using unfiltered entities and fancy prompting (proposed method, row 5, Table \ref{tab:Rad-QA}).

\textbf{RoBERTa v/s Benchmarks}: In row 4 (proposed method) of Table \ref{tab:Rad-QA}, RoBERTa, outperforms Bio/Sci-BERT and LUKE on all metrics along with long-context models BigBird and Longformer. Interestingly, RoBERTa even beats BioClinicalBERT by 1.9\% EM and 1.1\% F1, which was trained using much higher quality clinical notes. 

\textbf{Combined Prompting Styles}:
Here, we merge the contexts from both prompt styles (proposed method, rows 7 and 8, Table \ref{tab:Rad-QA}) for the filtered and the unfiltered entities separately. BERT shows better performance in row 7 i.e. using the unfiltered corpus (1.7\% EM increase over the filtered variant and roughly the same F1) and RoBERTa in row 8 i.e. using the filtered corpus (1.3\% EM and 0.8\%F1 over the unfiltered variant). Overall, this version of RoBERTa resulted in our best overall model on the validation data suggesting that incorporating a mixture of prompt styles will create more diverse corpora, enhancing domain alignment. 

\subsection{Investigating Information Leakage}
\label{app:information_leakage}

\begin{figure}[tbh]
    \centering
    \includegraphics[width=0.45\textwidth]{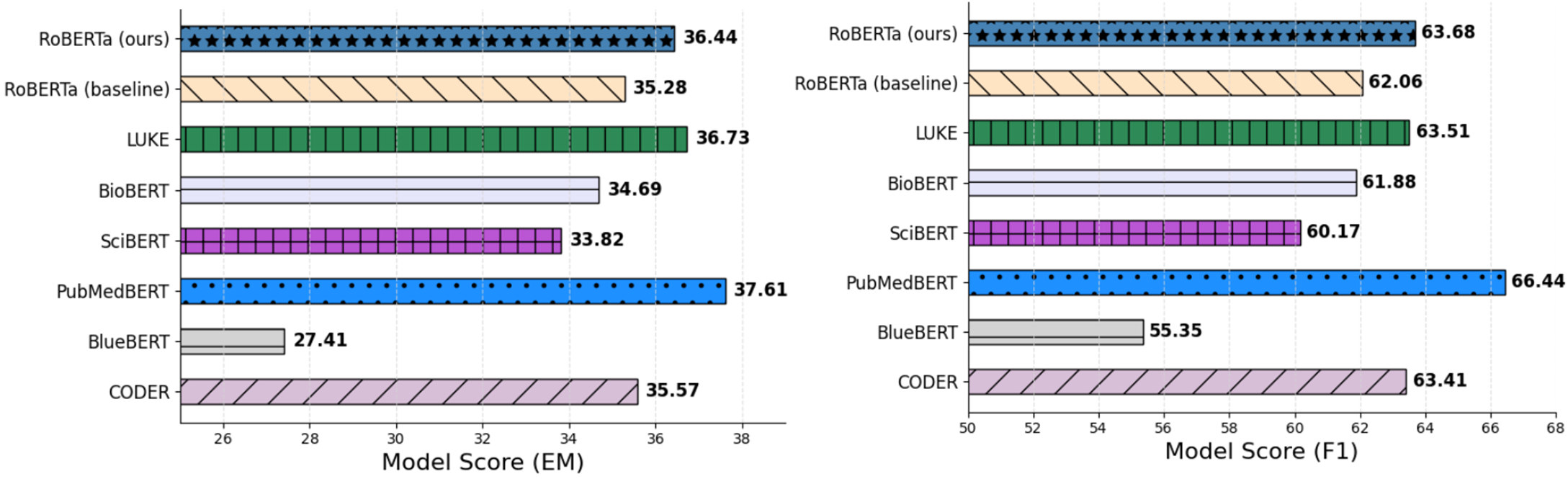}
    \caption{Information Leakage Validation Trials (Left - EM | Right - F1): RoBERTa (ours) was trained on a subset of the 47k corpus with entities only from the 80\% train set. All of the models were fine-tuned in the usual manner i.e. SQuAD$\rightarrow$COVID-QA (80\% train set) and evaluated on the 20\% test set.}
    \label{fig:modelscore}
\end{figure}


Given that the synthetic corpus generated for COVID-QA in \S \ref{sec:Exp} contains entities identified in the \textit{entire} COVID-QA dataset - not from the \textit{train} split within each \textit{fold} - we explore if the performance gains from \pretrainnew~are a result of information leak. To this end, we construct a roughly 80\%/20\% train/test split (1,676/343 records), ensuring no context overlap, and apply a suite of models to this new split. When applying our \pretrainnew, a synthetic corpus is generated \textit{only} from entities identified in the train split. 

RoBERTa subjected to \pretrainnew~still yields strong performance in this restricted scenario, only surpassed by PubMedBERT (and marginally by LUKE in EM) (Fig. \ref{fig:modelscore}) demonstrating that the improved performance on COVID-QA cannot be attributed to information leak from the test set. Although the scores are lower than Table \ref{tab:covidqa-all}, the relative scores produced by each model lead to a similar conclusion that \pretrainnew~yields optimal results.

\section{Conclusion}


We introduce \pretrainnew, an innovative pre-training solution to enhance the alignment of an LLM with rare-domain target problems. The distinctive feature of \pretrainnew~ lies in its automated synthesis of target-oriented data with the assistance of an LLM. Our experiments on \meqa~ demonstrate the effectiveness of \pretrainnew, shedding light on the limitations of widely used autoregressive LLMs.

\section*{Limitations}
We recognize that hallucination is still a concern with our method i.e. we do not provide a way to alleviate it. This is one area which our future work will focus on, i.e., developing ways to detect and remove hallucinated text from the generated corpus. In our analysis, we do not examine the \textit{quality} of the generated corpora and only rely on final task performance as an indicator of how good our generated corpus is. While we recognize the insights that we can gain from this analysis are interesting, we do not believe that its absence seriously undermines the proposed pipeline.

\section*{Ethics Statement}

As our work relied on publicly available datasets, we believe that the ethical ramifications here are limited. That being said, we recognize that to use RadQA, we had to acquire certifications to access it. This shows that even though the data in it is redacted, loosely disseminated patient reports are a threat to privacy. Moreover, we had to make sure that when generating our synthetic reports, we did not mention any patient names which even with a small probability might bear resemblance to an actual person.

\bibliography{custom}

\appendix

\section{Token Filtering}
We performed entity filtering as a common ablation technique for both datasets. We used regular expressions to remove entities with special characters such as *, !, etc., as well as specific text patterns like \texttt{https*} and \texttt{baby}. We implemented a length-based filter, retaining only entities longer than a certain number of characters. Additionally, for COVID-QA, we applied a second round of filtration using TF-IDF, considering the questions + context as the corpus and retaining the top 25k entities with the highest IDF scores. However, as this approach did not lead to substantial gains, we decided not to use it for RadQA. Due to the large number of possible combinations, we did not extensively explore these settings in our experiments.
\section{Synthetic Corpora Samples}
\label{app:corpora_samples}

\begin{figure*}
    \centering
    \frame{\includegraphics[width=\textwidth]{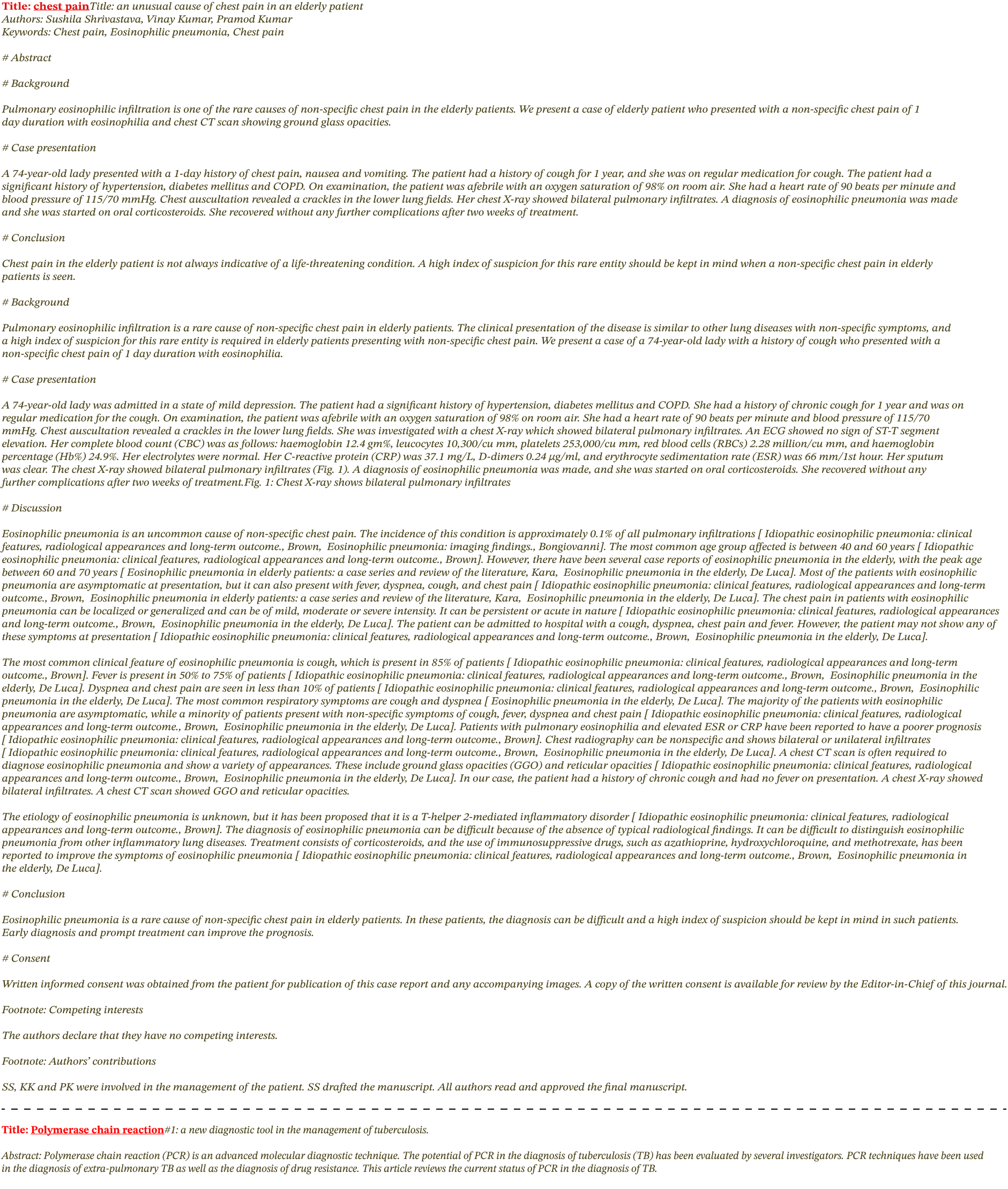}}
    \caption{\textbf{COVID-QA positive} examples (\underline{underline} = entity | \textcolor{red}{red} = prompt) | \textcolor{olive}{olive} = generation}
    \label{fig:covid-good}
\end{figure*}

\begin{figure*}
    \centering
    \frame{\includegraphics[width=\textwidth]{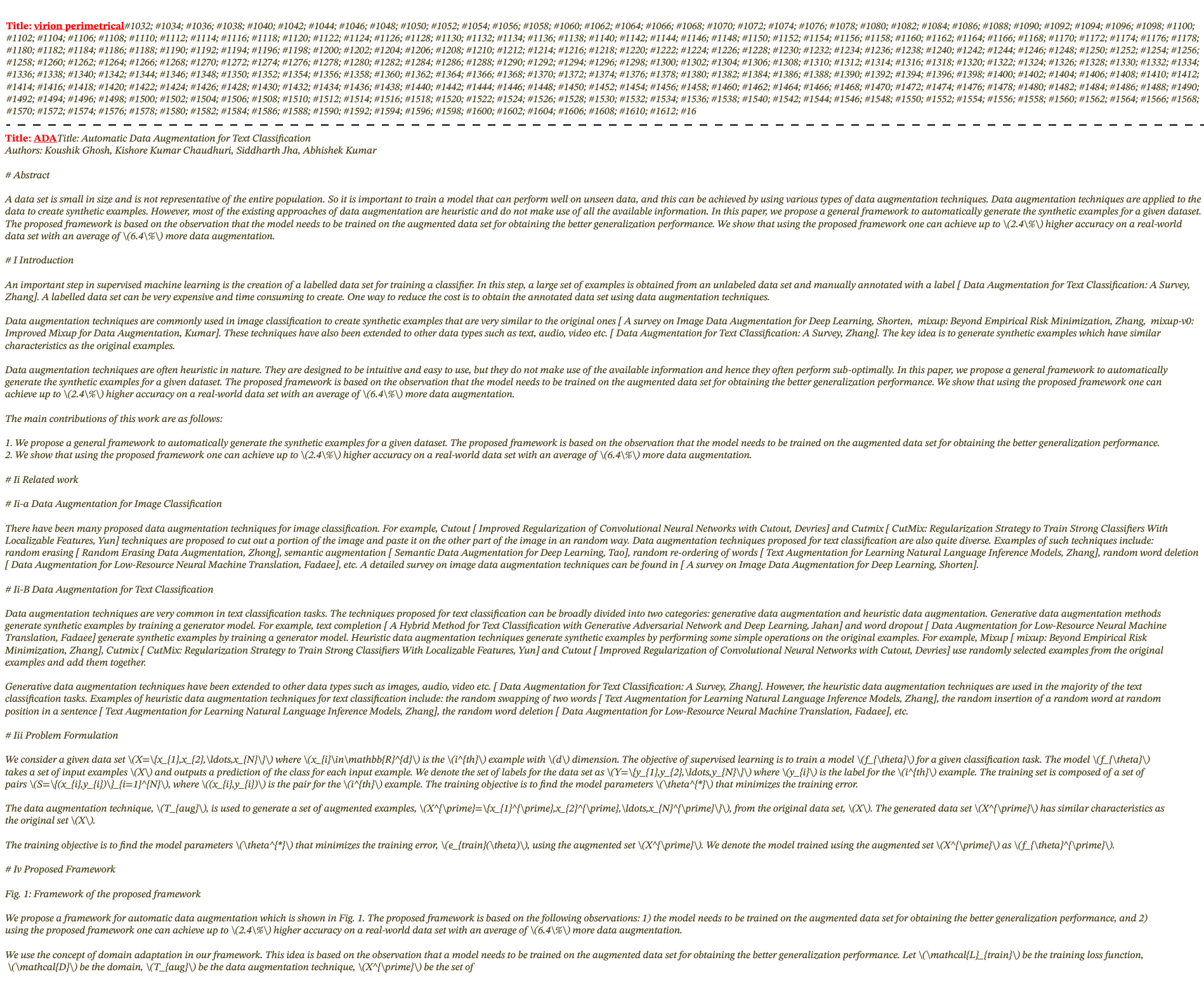}}
    \caption{\textbf{COVID-QA negative} examples (\underline{underline} = entity | \textcolor{red}{red} = prompt) | \textcolor{olive}{olive} = generation}
    \label{fig:covid-bad}
\end{figure*}

We provide samples from our generated corpus. In Figure \ref{fig:covid-good} we show two positive results for our COVID-QA-directed corpus. The top one is in the style of a research paper while the bottom one, though shorter in length, details useful information on the required entity. We show negative examples of the same in Figure \ref{fig:covid-bad}. The top one completely degrades into noise while the bottom one although coherent is talking about an unrelated topic.

\begin{figure*}
    \centering
    \frame{\includegraphics[width=\textwidth]{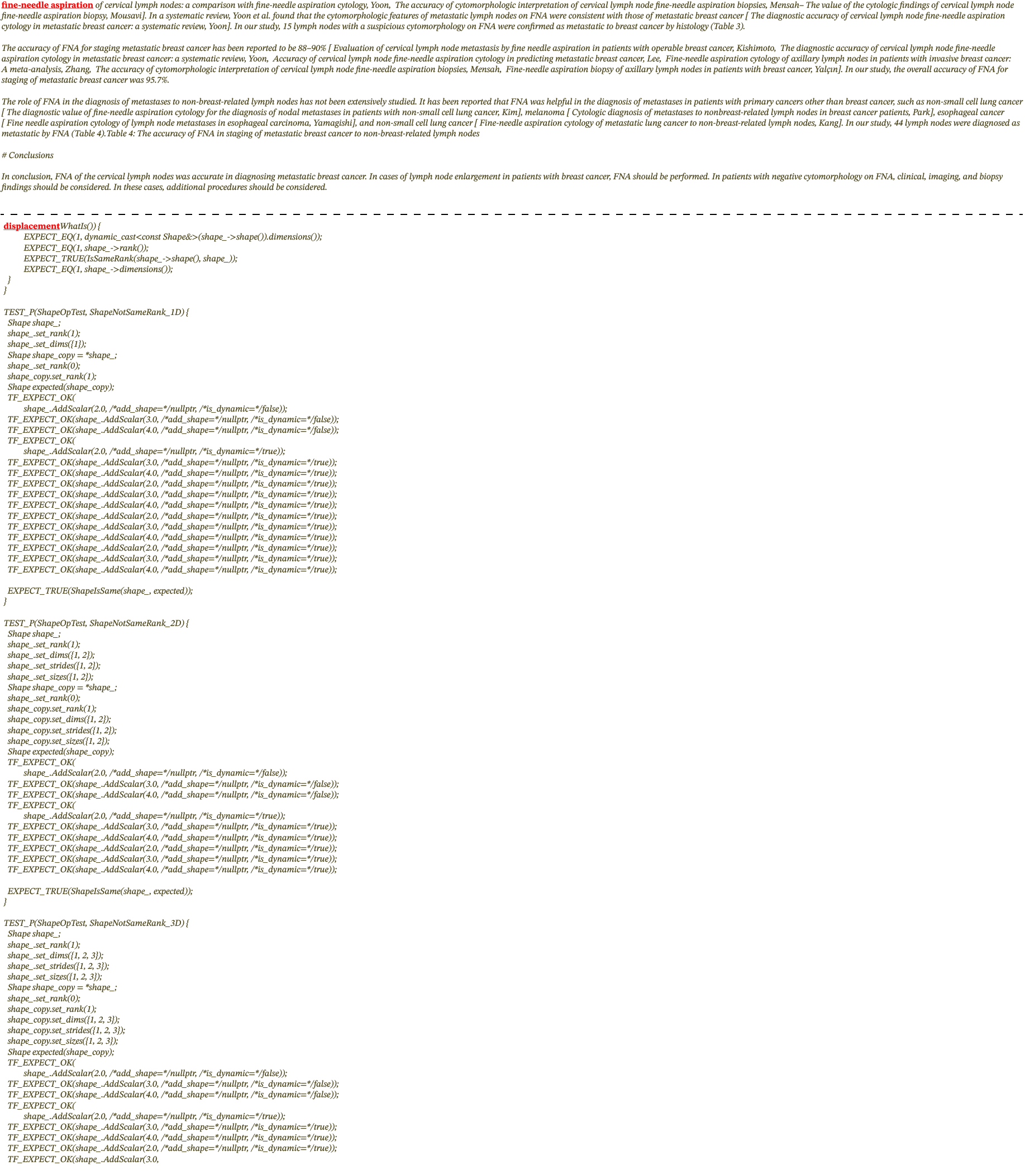}}
    \caption{\textbf{RadQA - normal prompt} (\underline{underline} = entity | \textcolor{red}{red} = prompt) | \textcolor{olive}{olive} = generation}
    \label{fig:radqa-normal}
\end{figure*}

\begin{figure*}
    \centering
    \frame{\includegraphics[width=\textwidth]{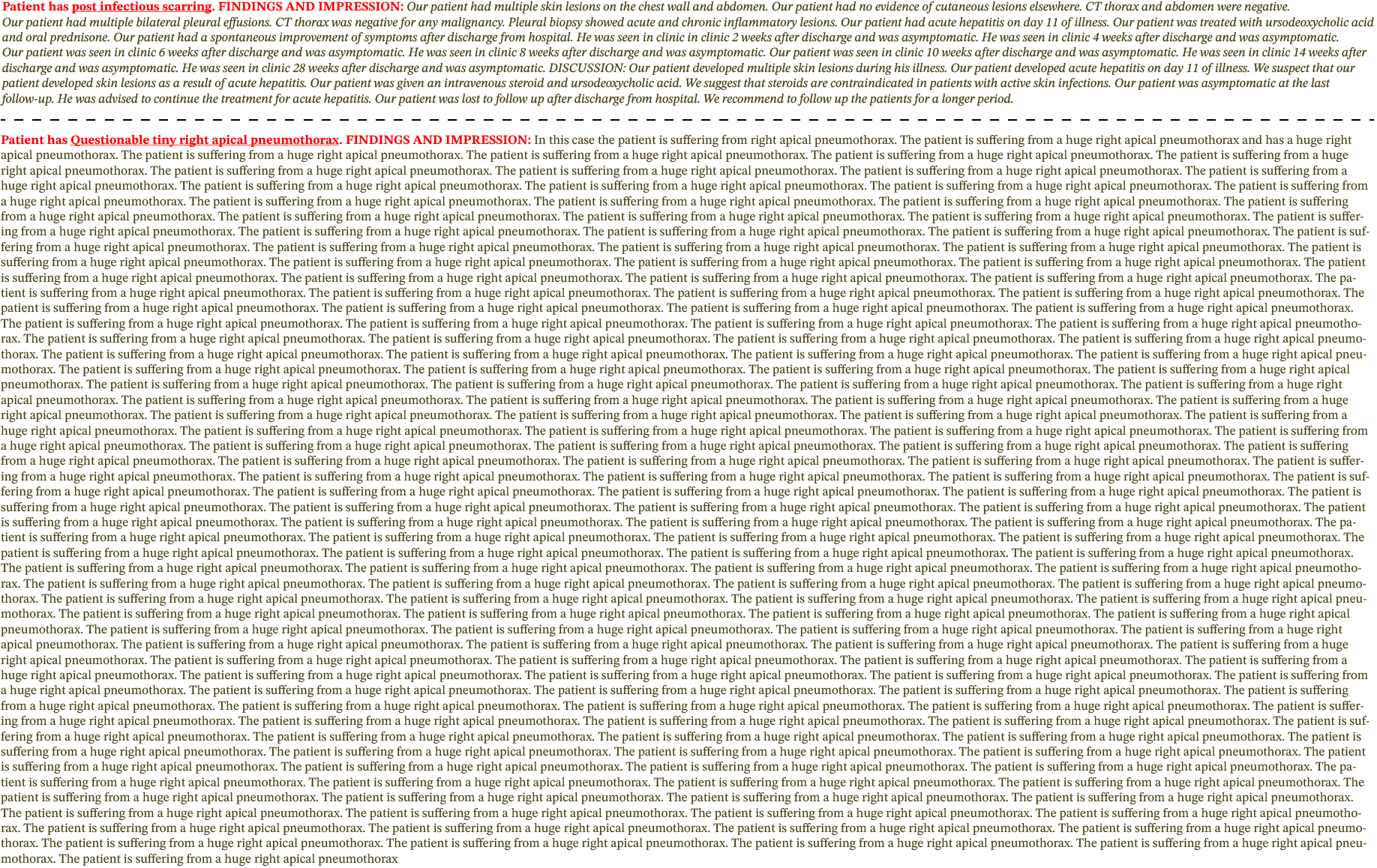}}
    \caption{\textbf{RadQA - fancy prompt} (\underline{underline} = entity | \textcolor{red}{red} = prompt) | \textcolor{olive}{olive} = generation}
    \label{fig:radqa-fancy}
\end{figure*}

For RadQA, we show similar examples. In Figure \ref{fig:radqa-normal} we show positive (top) and negative (bottom) samples from normal prompting. Note that since programming languages were a part of Galactica's training corpora, it sometimes generates the same during prompting. Finally, Figure \ref{fig:radqa-fancy} shows positive (top) and negative (bottom) examples of fancy prompting. While the positive sample shows a strong correlation with an actual radiology report, the bottom one completely breaks down.

\section{ChatGPT trials}
\label{app:chatgpt}

\begin{figure*}
    \centering
    \frame{\includegraphics[width=\textwidth]{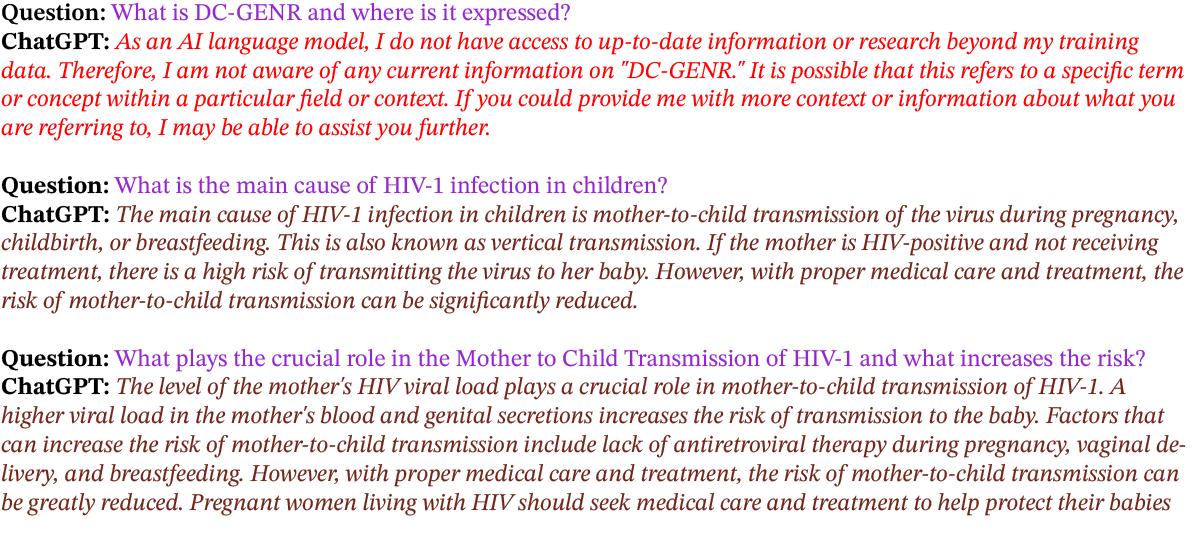}}
    \caption{Testing ChatGPT}
    \label{fig:chatgpt-res}
\end{figure*}

We were curious to see if ChatGPT (with GPT3.5) would be able to answer a few questions from COVID-QA. Only the questions were provided to it without the context since they were much longer than its input capacity. We did not test RadQA since its questions were heavily linked to the patient report on hand and were not as \textit{general} as the former's. Figure \ref{fig:chatgpt-res} shows a few examples from our trials. As we can see, it seems to answer the questions correctly. However, as mentioned before, these answers need validation from a professional in the field. Also, we see that sometimes it \textbf{cannot answer} questions on topics not found in its training data, a problem not faced by our encoder models which rely only on the provided context.

\section{Hardware Details}

To run our corpus generation code, we used a total of 8 40GB NVIDIA A100 or 8 32GB NVIDIA Tesla V100 cards, depending on our institute's job scheduler and availability. Although the former is a superior card, we preferred the latter since it gives a relatively quick turnaround and is a general-purpose GPU, unlike the A100 which is required for even heavier computing. We were requested by our system administrator to use the V100 over the A100, if it was possible, to avoid blocking off even more compute-heavy experiments, which we did not have an issue obeying. That being said, if the A100 cluster was free, we did take advantage of it to run our code. For pre-training and fine-tuning BERT/RoBERTa, we used only 1 A100 80GB GPU as RoBERTa was unable to fit on the 40/32 GB cards owing to the large batch size.

\section{Prompts Tried}

As mentioned before, we knew the prompt to be used for COVID-QA since it was a feature of Galactica. However, we had to experiment with a few prompt styles for RadQA before landing on our final one. We describe the prompts we tried below for the entity \texttt{small-bowel injury} and the reasons for not using them. The text in \texttt{courier} is the prompt and in \textit{italics} is the generated text.

\begin{itemize}
    \item \texttt{IMPRESSION: small-bowel injury} \textit{(SBI) was defined as injury to the small bowel wall, or transmural disruption of the bowel wall, in the absence of penetrating abdominal trauma ( The management of small-bowel injuries in the emergency room., Brown). Patients were excluded if the SBIs occurred after admission to the emergency department or were complicated by perforation. ...} (truncated)
    \begin{itemize}
        \item Issue - Did not include the \texttt{FINDINGS} section.
    \end{itemize}

    \item \texttt{FINDINGS: small-bowel injury} \textit{was observed in 11.1\% and 21.9\% of patients, respectively. Small-bowel injury associated with liver injury occurred in 10.1\%. The mortality rate was significantly higher in the small-bowel injury group than in the non-small-bowel injury group (11.6\% vs. 2.1\%, P = 0.005). This study shows that small-bowel injury is associated with a high mortality rate, and that this complication is an important cause of morbidity and mortality in patients with severe acute pancreatitis.}
    \begin{itemize}
        \item Issue - Did not include the \texttt{IMPRESSIONS} section.
    \end{itemize}

    \item \texttt{Patient has small-bowel injury. FINAL REPORT} \textit{This patient was treated with conservative methods. The small-bowel injury was treated surgically with resection of the bowel, primary anastomosis, and end-to-end anastomosis between the ends of the resected bowel. She did not have recurrence of intestinal obstruction after 3 years.}
    \begin{itemize}
        \item Issue - Did not include either \texttt{IMPRESSIONS} or \texttt{FINDINGS} section.
    \end{itemize}

    \item \texttt{John Doe has small-bowel injury. IMPRESSIONS} \textit{ : This case study was made possible by his support by a research grant from the National Institutes of Health (CA 10644).}
    \begin{itemize}
        \item Issue - Had to make sure that the pronouns used were same throughout the report otherwise it would not be consistent.
    \end{itemize}

    \item \texttt{Jane Doe has small-bowel injury. IMPRESSIONS} \textit{: This case study was made possible by the support of a research grant from the Department of Veterans Affairs.}
    \begin{itemize}
        \item Issue - Same as above.
    \end{itemize}

    \item \texttt{Patient I has small-bowel injury} \textit{He has no history of abdominal trauma or surgery. At first, he has only mild abdominal pain. But with the progress of disease, he has become distressed. At the initial examination, the abdominal pain was alleviated, but the patient still has slight distention of abdomen. He was admitted to our hospital on the third day of admission. His vital signs were: body temperature 37.5 °C, blood pressure 110/70 mmHg, pulse 105 beats/min, respiratory rate 26 breaths/min and oxygen saturation 97\% on room air. He had mild abdominal distention. The results of routine blood tests, urinalysis and abdominal ultrasound examination were normal. X-ray showed free air in the abdomen. ...} (truncated)
    \begin{itemize}
        \item Issue - Galactica could get confused between I (alphabet) and I (roman numeral).
    \end{itemize}

    \item \texttt{Patient A has small-bowel injury} \textit{ (Fig. 1). He has no history of abdominal trauma or surgery. At first, an abdominal CT scan was performed to evaluate abdominal pain. Abdominal CT showed a partial small-bowel obstruction at the splenic flexure of the colon (Fig. 2). An oral contrast medium was then administered via a nasogastric tube and an abdominal CT scan was performed. ...} (truncated)
    \begin{itemize}
        \item Issue - Galactica could get confused misinterpret ``A'' for the beginning of a sentence (it was observed for a few cases)
    \end{itemize}

    \item \texttt{Patient alpha has small-bowel injury} \textit{ requiring an open operation and was diagnosed with C. difficile colitis. The patient had a history of severe anemia and leukopenia, which were also observed in patient beta. Patient alpha was treated with a single course of intravenous moxifloxacin 400 mg/day, which resulted in the resolution of the diarrhea. He was followed closely after discharge. In patient beta, he had a similar episode of diarrhea after hospital admission, but the diarrhea did not resolve. He was prescribed with an oral vancomycin 1000 mg twice a day, and the diarrhea was resolved. ...} (truncated)
    \begin{itemize}
        \item Issue - This would have been a good choice. However, as we see above, patient qualifiers such as \texttt{A}, \texttt{I} and \texttt{alpha}, \textit{might} bias the model towards more \textit{male patients} (He). As such, we decided to drop the qualifier altogether.
    \end{itemize}

\end{itemize}

\section{Hyperparameters Used}
\label{app:hyperparams}

\begin{table}[ht]
\centering
\resizebox{\columnwidth}{!}{%
\begin{tabular}{l|l}
\textbf{Experiment} & \textbf{Hyperparameters}                                                                   \\ \hline
Corpus Generation &
  \begin{tabular}[c]{@{}l@{}}random seed: 42 \\ renormalize\_logits: True\\ do\_sample: True \\ max\_length (prompt + generated tokens): 2,048\\ top\_p: 0.9\\ temperature: 0.9\end{tabular} \\ \hline
Pre-Training        & \begin{tabular}[c]{@{}l@{}}batch\_size: 40\\ learning\_rate: 5e-5\\ epochs: 3\end{tabular} \\ \hline
Fine-Tuning (SQuAD) &
  \begin{tabular}[c]{@{}l@{}}batch\_size: 16\\ max\_input\_length (question + context): 384\\ stride: 128\\ learning\_rate: 2e-5\\ epochs: 3\\ n\_best (top n answer spans): 20\\ max\_answer\_length: 30\\ optimizer\_type: AdamW\end{tabular} \\ \hline
Fine-Tuning (COVID-QA) &
  \begin{tabular}[c]{@{}l@{}}batch\_size: 40\\ max\_input\_length (question + context): 384\\ stride: 128\\ learning\_rate: 2e-5\\ epochs: 1\\ n\_best (top n answer spans): 20\\ max\_answer\_length: 1000\\ optimizer\_type: AdamW\end{tabular} \\ \hline
Fine-Tuning (RadQA) &
  \begin{tabular}[c]{@{}l@{}}batch\_size: 16\\ max\_length: 384\\ stride: 128\\ learning\_rate: 3e-5\\ epochs: 1\\ n\_best (top n answer spans): 20\\ max\_answer\_length: 1000\\ optimizer\_type: AdamW\end{tabular}
\end{tabular}%
}
\caption{Hyperparameters for each experiment. We use three random seeds during pre-training and fine-tuning, \texttt{41,42,43} but only \texttt{42} when generating the corpus. This is done since otherwise running the entire pipeline from generation to training across all ablations would take an infeasible amount of time.}
\label{tab:hyperparams}
\end{table}

Hyperparameters for each experiment are detailed in Table \ref{tab:hyperparams}. These were selected mostly from preexisting implementations or through minimal exploration of known settings.

\section{Note on Stability}

All of our experiments were run using \texttt{PyTorch 1.13.1} and \texttt{Huggingface 4.26.1}. However, we have noticed fluctuations in results when training with other versions of these libraries. Thus, to replicate our scores to the best extent, we recommend installing the aforementioned versions of the packages. 


\section{Responses to Reviews}

In this section, we provide clarifying statements to questions raised during the review process. 

\begin{itemize}
    \item \textit{The generative models used in the evaluation (Galactia, MedLLaMMA, and MedAlpaca) do not appear to be producing valid outputs. One possible cause of this is the lack of instruction following ability in the models, it may be necessary to try instruction tuning or few-shot prompting that are suited to the assumed task or to try more capable recent generative models.}

    \begin{itemize}
        \item While not being able to produce valid outputs is a legitimate concern, we point out that EQA systems are evaluated using EM and F1, both of which count token overlap with the gold answer. As these models are free to produce text in an unconstrained manner, their answers display poor alignment with the gold reference. We observe that this is more often the case than producing invalid responses.

        \item Considering the complexity of the task, it would make sense to try instruction tuning, i.e., instead of span-extraction, make the objective span generation. That said, span generation is typically used for open-ended QA, i.e., abstractive QA instead of EQA. Additionally, we have not found studies in the existing literature that approach EQA in this manner. However, we agree that this will be an interesting experiment for the project. As for trying more capable LLMs, we tried ChatGPT (3.5) on a few samples as shown in Appendix \ref{app:chatgpt}. While we did not provide a context in those trials, we estimate using OAI documentation that processing the entire COVID-QA dataset would require upwards of \$450, which is quite expensive.
    \end{itemize}

    \item \textit{In your trial, how much does the proposed method depend on the prompts? Have you noticed any limitations in the prompt design?}

    \begin{itemize}
        \item This is a key point to the setup. The goal is to generate data that matches the content \& style of the downstream dataset. As such, writing prompts to obey this observation is very important. The prompt for COVID-QA was decided based on Galactica’s pre-training instructions. RadQA’s prompt required data examination to determine important features such as “Findings” and “Impressions”. Determining how much the overall method is impacted by prompt choice is challenging as it requires us to rerun the pipeline several times for corpora generated using different prompts. As such, our goal was to first write a prompt to mimic the downstream dataset’s contexts as best as possible, before generation and training.
    \end{itemize}

    \item \textit{I would also have liked to have had even a brief indication of the results that can be obtained in a specialist field other than medicine.}

    \begin{itemize}
        \item Although in this paper, we have shown a method to develop corpora for the medical domain using dataset-specific entities, there is no reason why the same cannot be applied to other closed domains such as legal or finance. This ideally will involve studying the dataset to gain insights on the linguistic characteristics to be mimicked by the synthetic data and using an associated entity recognizer that can perform better than general domain ones to better identify important entities.
    \end{itemize}

    \item \textit{One of the first issues addressed in the abstract and the introduction is the challenge of potential hallucinations of autoregressive LLMs. The pre-training framework TOP-Training is proposed "to address these challenges". However, the hallucination issue is neither addressed nor evaluated in the paper. The authors are aware of this problem and mention it as a limitation of the paper. While this is honest and appreciated, the objectives, abstract and introduction need to be more clarified.}

    \begin{itemize}
        \item The main point we highlight is that “hallucination” is a point of weakness in LLMs for solving EQA. However, we can still utilize their generation capabilities to produce synthetic “pre-training” data. The point on hallucination we mention under “Limitations” is that we do not implement a strategy to detect/filter such instances in the pre-training corpus as opposed to the task generations, when we apply the LLMs directly for the task.
    \end{itemize}

    \item \textit{Regarding the limitations of autoregressive LLMs for the EQA task, as written in the paper, their capabilities with EQA datasets are unexplored, and while one of the possible contributions of the paper was to show their limitations, autoregressive models are not evaluated in depth, which is normal given the scope of the article. The explanation of why they are not effective should be a bit more explicit.}

    \begin{itemize}
        \item As we’ve explained throughout the paper, such as the introduction, related work and benchmarking on both datasets (Table \ref{tab:covidqa-all}, \ref{tab:Rad-QA}), autoregressive LLMs are suboptimal for EQA. Apart from hallucination, another reason is the way EQA systems are evaluated, i.e., EM and F1, both of which require token overlap with the gold reference. As these models generate text in an open-ended manner, their performance drastically takes a hit.
    \end{itemize}

    \item \textit{\pretrainnew~ is a complex framework involving several important steps such as data argumentation and double fine-tuning with an open and specific domain dataset. This complexity makes the reasons why the method works less clear. Is it the combination of data augmentation and double fine-tuning that is successful, or would just fine-tuning in a specific domain dataset be enough, like is done in \cite{seo2024retrieval}? How does the use of different NER systems affect the results? etc.}

    \begin{itemize}
        \item \pretrainnew~refers to the entire pipeline of entity extraction + data generation + model training. While there are several steps, the complexity of our framework is mitigated by the fact that each step is straightforward and has a reason.~First, double pre-training is needed to align the open-domain model with the target domain.~Next, the two rounds of fine-tuning are needed to first learn the task on a high-quality dataset with several samples before approaching the target dataset with far less samples (both COVID-QA and RadQA have $\sim$8K samples $<<$ than SQuAD's 100K samples). We also find that directly fine-tuning models on the target dataset, i.e., without the first round of fine-tuning, leads to much worse performance as the randomly initialized span-extraction head is unable to learn from those few samples.~The impact of different NER tools is a bit more straightforward in that using a domain-aligned entity recognizer will lead to a higher number of quality identifications. That said, we did initially try Stanford’s Stanza \cite{qi-etal-2020-stanza}, but found Spacy’s extractions to be qualitatively better and thus decided to use it.
    \end{itemize}

\end{itemize}

\section{Model Cards}
\label{app:m_cards}

\begin{table}[htbp]
\centering
\resizebox{\columnwidth}{!}{\begin{tabular}{c|c}
\textbf{Model}                   & \textbf{Model Card (URL)}                                                                                              \\ \hline
BERT-Base, Cased                 & \href{https://huggingface.co/bert-base-cased}{bert-base-cased}                                                         \\
BERT-Base, Cased, SQuAD v1       & \href{https://huggingface.co/batterydata/bert-base-cased-squad-}{batterydata/bert-base-cased-squad-v1}                 \\
BERT-Base, Cased, SQuAD v2       & \href{https://huggingface.co/deepset/bert-base-cased-squad2}{deepset/bert-base-cased-squad2}                           \\
RoBERTa-Base                     & \href{https://huggingface.co/roberta-base}{roberta-base}                                                               \\
RoBERTa-Base, SQuAD v1           & \href{https://huggingface.co/csarron/roberta-base-squad-v1}{csarron/roberta-base-squad-v1}                             \\
RoBERTa-Base, SQuAD v2           & \href{https://huggingface.co/deepset/roberta-base-squad2}{deepset/roberta-base-squad2}                                 \\
BioBERT                          & \href{https://huggingface.co/dmis-lab/biobert-base-cased-v1.2}{dmis-lab/biobert-base-cased-v1.2}                       \\
SciBERT                          & \href{https://huggingface.co/allenai/scibert_scivocab_uncased}{allenai/scibert\_scivocab\_uncased}                       \\
SciBERT (+CORD-19)               & \href{https://huggingface.co/lordtt13/COVID-SciBERT}{lordtt13/COVID-SciBERT}                                           \\
PubMedBERT & \href{https://huggingface.co/microsoft/BiomedNLP-PubMedBERT-base-uncased-abstract-fulltext}{microsoft/BiomedNLP-PubMedBERT-base-uncased-abstract-fulltext} \\
BlueBERT   & \href{https://huggingface.co/bionlp/bluebert_pubmed_mimic_uncased_L-12_H-768_A-12}{bionlp/bluebert\_pubmed\_mimic\_uncased\_L-12\_H-768\_A-12}                   \\
CODER                            & \href{https://huggingface.co/GanjinZero/UMLSBert_ENG}{GanjinZero/UMLSBert\_ENG}                                         \\
LUKE                             & \href{https://huggingface.co/studio-ousia/luke-base}{studio-ousia/luke-base}                                           \\
XLNet, SQuAD v1                  & \href{https://huggingface.co/arrafmousa/xlnet-base-cased-finetuned-squad}{arrafmousa/xlnet-base-cased-finetuned-squad} \\
{\color[HTML]{FE0000} STonKGs *} & {\color[HTML]{FE0000} \href{https://huggingface.co/stonkgs/stonkgs-150k}{stonkgs/stonkgs-150k}}                        \\
RadBERT                          & \href{https://huggingface.co/zzxslp/RadBERT-RoBERTa-4m}{zzxslp/RadBERT-RoBERTa-4m}                                     \\
Clinical BERT                    & \href{https://huggingface.co/emilyalsentzer/Bio_ClinicalBERT}{emilyalsentzer/Bio\_ClinicalBERT}                         \\
BioMed-RoBERTa                   & \href{https://huggingface.co/allenai/biomed_roberta_base}{allenai/biomed\_roberta\_base} \\
MedLLaMA                         & \href{https://huggingface.co/chaoyi-wu/MedLLaMA_13B}{chaoyi-wu/MedLLaMA\_13B} \\
MedAlpaca                        & \href{https://huggingface.co/medalpaca/medalpaca-13b}{medalpaca/medalpaca-13b} \\
Galactica                        & \href{https://huggingface.co/facebook/galactica-1.3b}{facebook/galactica-1.3b} \\
Longformer, SQuAD v1                       & \href{https://huggingface.co/valhalla/longformer-base-4096-finetuned-squadv1}{valhalla/longformer-base-4096-finetuned-squadv1} \\
Longformer, SQuAD v2                       & \href{https://huggingface.co/mrm8488/longformer-base-4096-finetuned-squadv2}{mrm8488/longformer-base-4096-finetuned-squadv2} \\
BigBird, SQuAD v1                          & \href{https://huggingface.co/FredNajjar/NF-bigbird-squad}{FredNajjar/NF-bigbird-squad} \\
BigBird, SQuAD v2                          & \href{https://huggingface.co/FredNajjar/bigbird-QA-squad_v2}{FredNajjar/bigbird-QA-squad\_v2}
\end{tabular}
}
\caption{Model cards and URLs for all models used in our paper. * We wanted to use STonKGs \cite{balabin2022stonkgs}. However, there was no vocabulary file for the model which resulted in errors.}
\label{tab:mod_cards}
\end{table}

All models used in this study were downloaded from HuggingFace \footnote{\url{https://huggingface.co/}}. Each model, along with its model card (name as it appears in the HuggingFace model hub) and URL is listed in Table \ref{tab:mod_cards}.

\end{document}